%% file: main_arxiv.tex
\documentclass{article}
\usepackage{ijcai25}

\usepackage{times}
\usepackage{soul}
\usepackage{url}
\usepackage[hidelinks]{hyperref}
\usepackage[utf8]{inputenc}
\usepackage[small]{caption}
\usepackage{graphicx}
\usepackage{amsmath}
\usepackage{amsthm}
\usepackage{booktabs}
\usepackage{algorithm}
\usepackage{algorithmic}
\usepackage[switch]{lineno}
\usepackage{dsfont}
\usepackage{multirow}
\usepackage{amssymb}
\usepackage{makecell}
\usepackage{colortbl}
\usepackage{color,xcolor}
\usepackage{float}
\usepackage{marvosym}
\usepackage{ifsym}

\title{Accelerating Diffusion-based Super-Resolution with \\ Dynamic Time-Spatial Sampling}

\newcommand{\modified}[1]{#1}

\definecolor{green}{RGB}{0,127,0}
\author{
\modified{
Rui Qin$^{1, \dag}$\and
Qijie Wang$^{1, \dag}$\and
Ming Sun$^2$\and
Haowei Zhu$^1$\and
Chao Zhou$^2$\And
Bin Wang$^{1, \text{\Letter}}$\\
\affiliations
\modified{$^1$School of Software, Tsinghua University, Beijing, China\\
$^2$Kuaishou Technology, Beijing, China\\}
\emails
qr20@mails.tsinghua.edu.cn,
wqj24@mails.tsinghua.edu.cn,
sunming03@kuaishou.com, \\
zhuhw23@mails.tsinghua.edu.cn,
zhouchao@kuaishou.com,
wangbins@tsinghua.edu.cn
}
}

\newcommand{\best}[1]{\underline{#1}}

\begin{document}

\maketitle

\let\thefootnote\relax

\input{data/abstract.tex}

\input{data/intro.tex}

\input{data/related.tex}

\input{data/method.tex}

\input{data/experiment.tex}

\input{data/conclusion.tex}

\clearpage
\section*{\modified{Acknowledgements}}
\modified{This work was supported by the National Natural Science Foundation of China under Grant 62072271.}

\section*{\modified{Contribution Statement}}
\modified{$\dag$ Co-first authors. \Letter\ Corresponding author.}

\appendix
\input{data/appendix.tex}

\clearpage
\bibliographystyle{named}
\bibliography{main}

\clearpage


\end{document}

%% file: data/abstract.tex

\begin{abstract}
    Diffusion models have gained attention for their success in modeling complex distributions, achieving impressive perceptual quality in SR tasks. However, existing diffusion-based SR methods often suffer from high computational costs, requiring numerous iterative steps for training and inference. 
    Existing acceleration techniques, such as distillation and solver optimization, are generally task-agnostic and do not fully leverage the specific characteristics of low-level tasks like super-resolution (SR).
    In this study, we analyze the frequency- and spatial-domain properties of diffusion-based SR methods, revealing key insights into the temporal and spatial dependencies of high-frequency signal recovery. Specifically, high-frequency details benefit from concentrated optimization during early and late diffusion iterations, while spatially textured regions demand adaptive denoising strategies. Building on these observations, we propose the Time-Spatial-aware Sampling strategy (TSS) for the acceleration of Diffusion SR without any extra training cost. TSS combines Time Dynamic Sampling (TDS), which allocates more iterations to refining textures, and Spatial Dynamic Sampling (SDS), which dynamically adjusts strategies based on image content. Extensive evaluations across multiple benchmarks demonstrate that TSS achieves state-of-the-art (SOTA) performance with significantly fewer iterations, improving MUSIQ scores by 0.2~$\sim$~3.0 and outperforming the current acceleration methods with only half the number of steps. 
\end{abstract}

%% file: data/intro.tex

\section{Introduction}

\begin{figure*}[t]
    \centering
    \includegraphics[width=\linewidth]{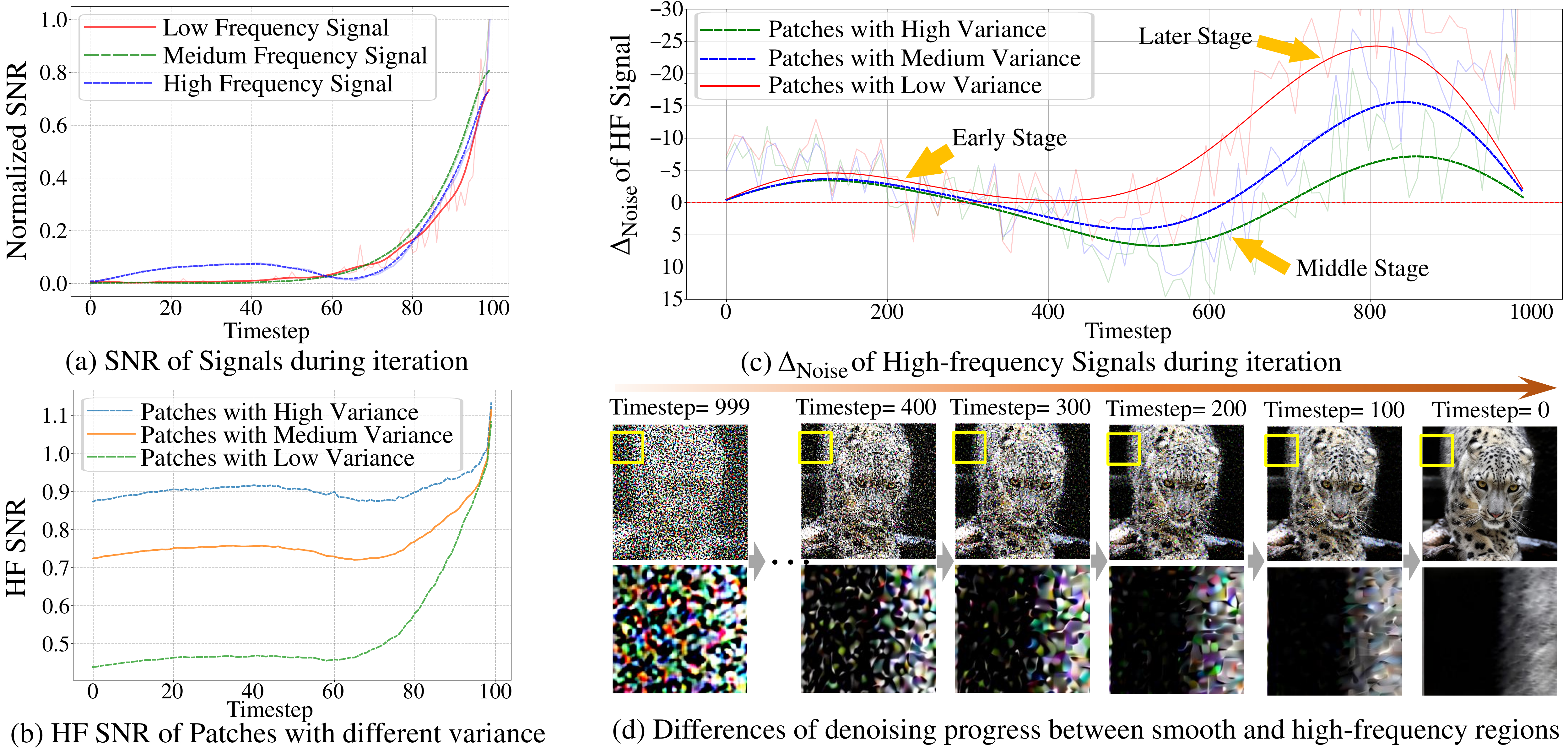}
    \caption{
    \modified{(a) SNR of different frequency components in SUPIR denoising, where high-frequency signals show a unique two-stage pattern.
    (b) SNR of high-frequency signals in different spatial regions during the denoising process of SUPIR.  
    (c) Noise amplitude in high-frequency regions of SUPIR denoising: higher variance shortens the positive optimization phase.
    (d) Denoising visualization of a RealPhoto60 sample.}}
    \label{fig:fig1}
    
\end{figure*}

Image super-resolution (SR)~\cite{realesrgan,bsrgan,swinir,qin2023rtcnet,liu2023rclut,zhao2023biqa,pbasr,bao2025plug} aims to reconstruct high-resolution (HR) images from low-resolution (LR) inputs. Recently, diffusion models~\cite{ho2020ddpm,song2020ddim} have gained attention for their ability to model complex distributions, achieving notable success in SR~\cite{chen2024cassr,wang2024sinsr,yang2025pasd,yu2024supir,wu2024seesr,qu2025xpsr}, particularly in perceptual quality~\cite{clipiqa,qalign}.
Diffusion-based image super-resolution methods take two primary approaches: integrating the low-resolution image into a task-specific denoiser~\cite{saharia2022sr3,stablesr} or adapting the reverse diffusion process of pre-trained models~\cite{wu2024seesr,yang2025pasd,yu2024supir}. These methods are computationally intensive, requiring 1000 steps for training and several, such as 20 (PASD~\cite{yang2025pasd}), 50 (SUPIR~\cite{yu2024supir}), or more steps (StableSR~\cite{stablesr}) during testing.

Efforts to accelerate denoising generation focus on sampler acceleration and distillation, achieving results in 10 or fewer steps~\cite{yue2024resshift,wang2024sinsr}. Most Diffusion SR methods~\cite{stablesr,yu2024supir,yang2025pasd} adopt these general strategies without considering the unique frequency characteristics of low-level vision tasks. However, in fact, recent studies like STAR~\cite{xie2025star} have highlighted the diverse recovery of diffusion-based SR across frequency domains, suggesting the potential to learn low- and high-frequency information at different training stages.
Despite these insights, these works primarily focus on modifying the training process and optimization losses. Given the availability of many large-scale open source and pre-trained diffusion SR models~\cite{yang2025pasd,yu2024supir,stablesr}, we aim to develop a tailored training-free acceleration strategy by leveraging the characteristics of Diffusion SR methods with the spatial and frequency information, seeking to enhance the performance of these existing models at minimal cost.

To analyze frequency-based disparities in the denoising process, we conduct a tiny experiment, using SUPIR, one of the latest typical state-of-the-art Diffusion SR methods, on the RealPhoto60 dataset~\cite{yu2024supir}. RealPhoto60 comprises 60 real-world images from common benchmarks. To explore the frequency characteristics, we applied Fourier transformation~\cite{fft} to categorize spectra into low, medium, and high-frequency signals. To observe the time domain dynamics, we recorded the signal-to-noise ratio (SNR) of intermediate and final outputs over the 100-step inference. As shown in Fig.~\ref{fig:fig1}.a, SNR improvements were most pronounced in the later stages across all frequency bands. Notably, \textbf{unlike low and medium frequency components,  high-frequency components uniquely exhibited visible SNR gains in the early stages}, indicating the critical role of early denoising in restoring high-frequency details.

Furthermore, we analyzed spatial variations in high-frequency recovery by cropping RealPhoto60 samples into 128-sized patches and categorizing them as smooth, medium-textured, or highly textured based on variance. We recorded the SNR of high-frequency signals for each category throughout the iterations (Fig.~\ref{fig:fig1}.b), and analyzed the noise magnitude changes (Fig.~\ref{fig:fig1}.c). For an intuitive comparison, Fig.~\ref{fig:fig1}.d illustrates the denoising process for a typical sample with both low- and high-frequency regions, showing that smooth areas recover early, while high-frequency regions, such as fur, recover significantly in the final $0 \sim 100$ steps. \textbf{Spatially, regions with more high-frequency information concentrate recovery in the initial and final stages, whereas content variation influences denoising dynamics.}

Based on the above observations, efficiently generating high-frequency details requires leveraging their unique temporal optimization, which is concentrated in the early and late iterations. Therefore, we propose a Time Dynamic Sampling (TDS) strategy that prioritizes high-frequency signal recovery and enhances texture perception by allocating more denoising steps to diffusion stages critical for refining high-frequency details.
Furthermore, from a spatial perspective, the sampling strategy must adapt to variations in image content. To achieve this, we introduce Spatial Dynamic Sampling (SDS), which dynamically adjusts the sampling frequency based on spatial content, ensuring alignment with the characteristics of different image regions.
By integrating these two strategies, we propose Time-Spatial-Aware Sampling (TSS), a novel framework to accelerate existing diffusion-based SR without additional training costs. As both strategies require no additional training and only minimal code modifications, TSS offers an efficient and broadly applicable solution.

Evaluations on six BSR benchmarks across various metrics illustrate that TSS significantly improves the performance of various diffusion SR methods within a few iterations, without incurring additional training costs. TSS consistently achieves an increase of 0.2 $\sim$ 3.0 of MUSIQ in diverse SR diffusion frameworks and datasets. Remarkably, TSS outperforms the current state-of-the-art method while using only half the number of steps. Our main contributions are as follows:
\begin{enumerate}
    \item We identified the temporal and spatial dynamics of diffusion-based methods in high-frequency detail recovery of image super-resolution tasks.

    \item Based on the observations, we propose the Time-Spatial-aware Sampling strategy (TSS) to achieve training-free acceleration for diffusion-based image super-resolution.
    
    \item  Comprehensive evaluations across multiple real-world SR benchmarks show that TSS achieves state-of-the-art performance with fewer denoising iterations.
\end{enumerate}

%% file: data/related.tex

\section{Related Work}
\subsection{Diffusion Model in Super-Resolution}
Due to their exceptional ability to generate high-quality images, diffusion-based super-resolution methods have garnered widespread attention. Early approaches leveraged low-resolution images as guidance by training a conditional DDPM~\cite{ho2020ddpm,kawar2022ddrm,saharia2022sr3}, or conditionally steering a pre-trained DDPM~\cite{choi2021ilvr} to perform super-resolution tasks. Recently, several studies have utilized pre-trained text-to-image (T2I) models, such as Stable Diffusion (SD)~\cite{rombach2022sd,podell2023sdxl}, harnessing learned priors to address super-resolution challenges and achieve high-quality image enhancements. These methods either train a ControlNet~\cite{zhang2023controlnet,yang2025pasd,yu2024supir,chen2024cassr} or an additional encoder that encodes guidance features~\cite{wu2024seesr,stablesr}, both of which have demonstrated outstanding performance. Yet, as mentioned earlier, these methods require a large number of diffusion steps, resulting in high computational costs.

\subsection{Diffusion Model Acceleration}
Recent state-of-the-art super-resolution methods still require dozens or even hundreds of diffusion steps, even with acceleration techniques like DDIM~\cite{song2020ddim}, leading to significant time overhead. Reducing diffusion steps often degrades output quality. Various pruning~\cite{zhu2024dipgo}, caching~\cite{ma2024deepcache}, and distillation~\cite{yin2024dmd} techniques have accelerated general T2I diffusion models while preserving generation quality.
In super-resolution, ResShift~\cite{yue2024resshift} reduces diffusion steps by modeling a Residual Shifting Markov chain between high-resolution and low-resolution images but requires \modified{a considerable cost of 500K iterations} training from scratch. 
Distillation approaches~\cite{wu2024osediff,he2024tad} have achieved one-step diffusion in super-resolution tasks, but their training costs remain high, and their performance is constrained by the teacher models. 
Therefore, given the training costs and limitations of existing methods, exploring a training-free acceleration framework for diffusion-based super-resolution is valuable and necessary.

%% file: data/method.tex

\section{Method}
 
\subsection{Revisiting Diffusion in Super-Resolution}
\label{sec:Revisiting}
To investigate the relationship between frequency signals, spatial characteristics, and temporal steps in denoising, we conducted two demo experiments using the representative SOTA method SUPIR~[\citeyear{yu2024supir}], a classic framework based on pre-trained SD~[\citeyear{rombach2022sd}] with fine-tuned ControlNet~[\citeyear{zhang2023controlnet}]. The experiments are conducted on the RealPhoto60 dataset containing 60 real-world images from benchmarks like RealSR~[\citeyear{realsr}], DRealSR~[\citeyear{drealsr}], and web sources.

\paragraph{Frequency Signal Analysis}
We analyze the relationship between denoising stages and frequency-specific signal recovery by computing the signal-to-noise ratio (SNR) over 100-step inference using the Fourier Transform, with the final denoised result as a noise-free reference and intermediate results decoded from feature representations.
As shown in Fig.~\ref{fig:fig1}.a, SNR increases significantly in the late stages (400 $\sim$ 0 steps). Notably, high-frequency (HF) components, unlike low and medium frequencies, also exhibit visible SNR improvement in the early stages (1000 $\sim$ 700 steps).
To further investigate HF signals crucial for SR, Fig.~\ref{fig:fig1}.c presents the noise delta throughout denoising, where noise represents the residual with the final result, and delta denotes stepwise changes. Noise decreases in the early (1000 $\sim$ 700) and late (400 $\sim$ 0) stages but rises in the middle, mirroring the SNR trend in Fig.~\ref{fig:fig1}.a. 
This suggests that, unlike low and medium frequencies, early denoising stages are also crucial for HF recovery, while intermediate stages may hinder optimization.
\paragraph{Spatial Dynamics Analysis}

To examine the spatial dynamics of denoising across regions with varying content, we analyzed signals in image patches of different textures. RealPhoto60 images were divided into non-overlapping 128×128 patches and classified as smooth, medium textured, or highly textured based on variance, applying Frequency Signal Analysis above to each category. 
As shown in Fig.~\ref{fig:fig1}.c, higher variance patches have a narrower range of time steps for effective optimization in the early and late stages. Fig.~\ref{fig:fig1}.d further illustrates that smooth regions undergo visible denoising earlier than fur-textured areas during late iterations. This suggests that denoising effectiveness varies across textures, with high-texture patches containing more high-frequency signals exhibiting shorter, more concentrated optimization stages in both early and late iterations.

In summary, the restoration of high-frequency (HF) signals is unevenly distributed across both the denoising process and spatial regions within an image. However, the widely used traditional uniform, data-independent sampling methods haven't taken the spatiotemporal dynamics of HF signal denoising into account, which are critical for super-resolution tasks. Additionally, acceleration strategies relying on distillation or pruning often incur additional training time. Consequently, we propose that an optimal acceleration strategy should meet the following criteria:
\begin{enumerate}
    \item \textbf{Cost-Effectiveness.} 
    The strategy should achieve superior performance with fewer iterations while keeping additional training costs minimal.
    \item \textbf{Generality.} The strategy should exhibit robust generalizability, enabling flexible integration across a wide range of established super-resolution frameworks.
\end{enumerate}

\subsection{Time-Spatial-aware Sampling}
\begin{figure}[b]
    \centering
    \includegraphics[width=\linewidth]{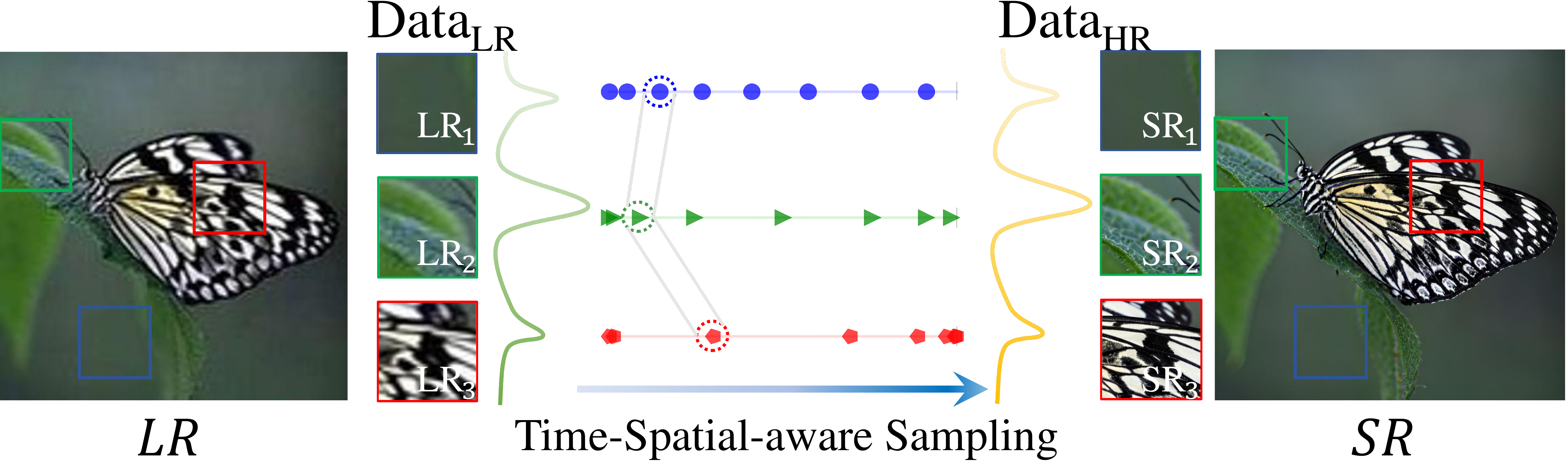}
    \caption{Overview of the proposed Time-Spatial-aware Sampling.}
    \label{fig:pipeline}
\end{figure}

To address these issues and meet the requirements, we propose Time-Spatial-aware Sampling (TSS), a training-free content-adaptive accelerated sampling strategy (in Fig.~\ref{fig:pipeline}). TSS integrates two core strategies: Time Dynamic Sampling and Spatial Dynamic Sampling, which are elaborated below.
\subsubsection{Time Dynamics Sampling (TDS)}
\begin{figure}[tbp]
    \centering
    \includegraphics[width=0.8\columnwidth]{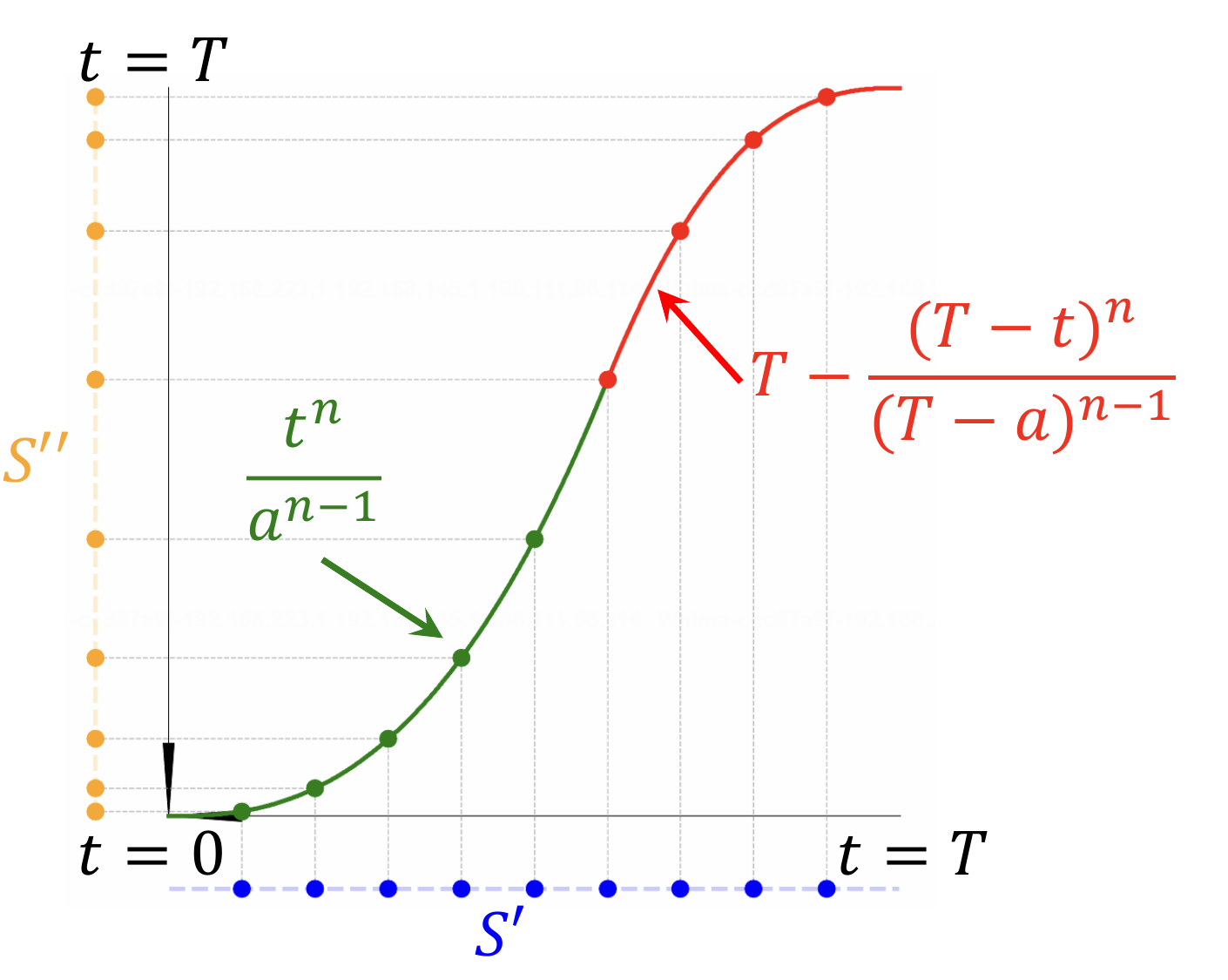}
    \caption{Illustration of the Time Dynamic Sampling strategy.}
    \label{fig:sampling}
\end{figure}

As discussed in Sec.~\ref{sec:Revisiting}, high-frequency signal optimization is concentrated in the early and late stages. Based on this observation, we propose the Time Dynamic Sampling (TDS) strategy with a non-uniform, adaptive timestep allocation. The non-uniform sampling design should meet two key criteria:
1) Higher sampling density in the early and late stages. 2) Adjustable non-uniformity, including uniform sampling as a special case.
To achieve this, we propose a non-uniform denoising approach focused on high-frequency information by implementing a non-uniform resampling strategy in the time-step schedule. Specifically, for a given number of training steps $T$ and training scheduler $S=\{ 1,2,3,... ,T\}$, the scheduler $S'$ for common uniform sampling of $T'$ steps is
\begin{equation} 
\label{eq:uniform}
S'(T,T')= \{t_k \| t_k = \lfloor k  \cdot \frac{T}{T'} \rfloor,  k \in \{1, 2, ..., T'\} \}.
\end{equation} 
To incorporate adaptive non-uniformity, we resample the uniform sampling scheduler using a two-stage polynomial function to derive the modified scheduler $S''$, which is defined as:
\begin{equation}
    \label{eq:power}
        S''(a, n, T, T') = \{ f(t, a, n, T, T') \| t \in S'(T,T') \}, 
\end{equation}
\begin{equation}
    \label{eq:non-uniform_sampling}
    f(t, a, n, T, T') = \begin{cases}
        \frac{t^n}{a^{n-1}} , t < a \\
        T - \frac{(T-t)^n}{(T-a)^{n-1}}, t \geq a \\
        \end{cases},
\end{equation} 
where $a$ denotes the transition point between early and late stages, and $n$ is the power factor. Fig.~\ref{fig:sampling} illustrates an example of the resampling function. When $n > 1$, sampling density is concentrated at $t = 0$ and $t = T$, while $n$ approaches 1, the function gradually converges to uniform sampling. Compared to uniform sampling, TDS enhances high-frequency signal recovery by prioritizing early and late-stage sampling, which is crucial for high-frequency restoration. 
\subsubsection{Spatial Dynamic Sampling (SDS)}
TDS improves high-frequency signal restoration at the overall image level. However, as discussed in the previous section, different image regions exhibit distinct denoising dynamics, requiring region-specific sampling within a single image. To address this, we introduce Spatial Dynamic Sampling (SDS), which extends TDS with  Variance-Adaptive Smooth Sampling and Spatial Dynamic Time Embedding.

\begin{figure}[tbp]
    \centering
    \includegraphics[width=0.95\columnwidth]{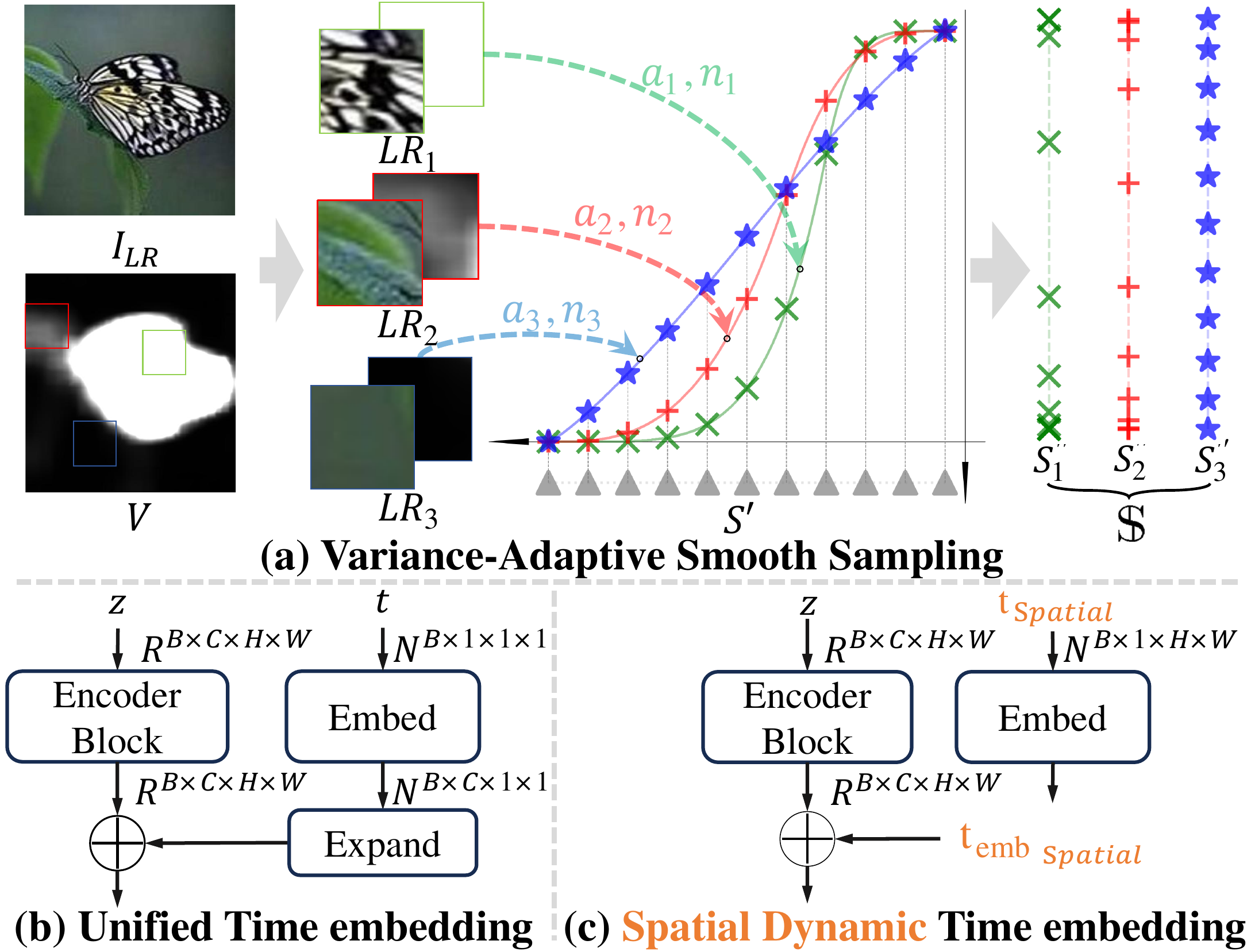}
    \caption{(a) The Variance Adaptive Smooth Sampling strategy. (b) The common unified embedding injection strategy. (c) The proposed Spatial Dynamic Time Embedding injection strategy.}
    \label{fig:spatial_embedding}
\end{figure}

\noindent \textit{Variance-Adaptive Smooth Sampling} 
To achieve dynamic scheduling across different image regions, as illustrated in Fig.~\ref{fig:spatial_embedding}.a, we first compute the local variance $V_0\in R^{H\times W}$ of the image $I_{LR} \in R^{H\times W \times 3}$ on grayscale using a $33 \times 33$ field around each pixel and post-process it with Gaussian blurring of the same size and min-max normalization for smoothness:
\begin{equation}
    \label{eq:gaussian}
    V_g = \mathrm{norm}(\mathrm{GaussianBlur}(V_0)).
\end{equation}
Subsequently, we derive the non-uniform sampling strategy for each position based on the local smoothed variance $V_g$. Specifically, we assign an independent time scheduler to each position and adaptively adjust it based on local variations. The overall spatial-adaptive time scheduler $\mathds{S}$ is defined as the set of schedulers for each position:
\begin{equation}
    \label{eq:spatial}
    \mathds{S} = \{ S''(a(v_{g_{i,j}}), n(v_{v_{g_{i,j}}}), T, T') \| v_{g_{i,j}} \in V_g\},
\end{equation}
where $v_{g_{i,j}}$ denotes the local variance at position $(i,j)$ which controls $a$ and $n$. The spatial timestep $t_{spatial_k}$ for the $k^{th}$ denoising iteration represents a set of timesteps for each pixel, 
\begin{equation}
    \label{eq:spatial_time}
    t_{spatial_k} = \{ t_k \in \mathds{S}_{i,j} | i \in [0,H], j \in [0,W] \},
\end{equation}
where $\mathds{S}_{i,j}$ denotes the single scheduler in $\mathds{S}$ for position $(i,j)$.
Considering computational efficiency and simplicity of implementation, we use linear functions for the projection $a()$ and $n()$, which are defined as follows:
\begin{equation}
    \label{eq:projection}
    n(v) = v(N_{\text{max}}-N_{\text{min}})+N_{\text{min}}, a(v) = v(A_{\text{max}}-A_{\text{min}})+A_{\text{min}},
\end{equation}
where $N_{\text{max}}$, $N_{\text{min}}$, $A_{\text{max}}$, and $A_{\text{min}}$ denote the range of $n$ and $a$.
As shown in Fig.~\ref{fig:spatial_embedding}.a, we assign different $a$ and $n$ values based on the local variance.
In this way, we ensure that high-frequency texture regions with high variance (red block in Fig.~\ref{fig:spatial_embedding}.a) undergo greater sampling inhomogeneity, concentrating steps in the early and late phases to enhance high-frequency recovery. Conversely, smooth regions with lower variance (blue block in Fig.~\ref{fig:spatial_embedding}.a) are sampled more uniformly, facilitating the restoration of smooth areas.

\noindent \textit{Spatial Dynamics Time Embedding}
While Variance-Adaptive Smooth Sampling allows for location-specific timestep allocation, segmenting images into separate patches to accommodate spatially varying timesteps will increase processing time and cause boundary artifacts.
To adapt pre-trained denoising networks for spatial timestep embeddings in one step, we introduce a simple yet effective spatiotemporal embedding injection strategy. Specifically, existing methods generate individual timestep embedding $t_{emb} \in R^{C}$ and add them to main branch features $z \in R^{B\times C \times H \times W}$ at all spatial locations via automatic expansion (Fig.~\ref{fig:spatial_embedding}.b), which can be formulated as
\begin{equation}
    \label{eq:embedding}
    t_{emb} = \mathrm{Emb}(t),z = z + \mathrm{expand}(t_{emb}),
\end{equation}
where the expanded version $ \mathrm{expand}(t_{emb})$ matches the spatial dimensions of the main branch features $z$. 
In contrast, our spatial timestep embedding is the set of embeddings for each spatial location in $t_{spatial}$, which has the same spatial dimensions as the main branch features $z$, as depicted in Fig.~\ref{fig:spatial_embedding}.c Thus, it can be directly added to the main branch features at all spatial locations, which can be expressed as:
\begin{equation}
    \label{eq:spatial_embedding}
    t_{emb_{spatial}} = \{\mathrm{Emb}(t_{i,j}) | t_{i,j} \in t_{spatial}\},
    z = z + t_{emb_{spatial}}.
\end{equation}
This allows the network to adapt to spatially varying timestep embeddings while integrating flexibly into existing architectures without extra cropping or multiple forward passes.

%% file: data/experiment.tex

\section{Experiment}

\subsection{Experiment Setup}
\paragraph{Implementation Details}

\modified{Our approach modifies only the sampling strategy during inference, requiring no extra training. Experiments were conducted on an NVIDIA A800 GPU (80GB) using the official codebases and pre-trained weights. The PyTorch framework was used for implementation, and the detailed hyperparameters are provided in the full version.}

\paragraph{Testing Datasets}

\modified{Performance was evaluated on both synthetic and real-world data. Synthetic data used DIV2K~[\citeyear{div2k}] with LR images degraded by BSRGAN~[\citeyear{bsrgan}]. Real-world evaluations used DRealSR~[\citeyear{drealsr}], RealPhoto60~[\citeyear{yu2024supir}], and RealSR~[\citeyear{realsr}], along with face SR benchmarks WebPhoto-Test~[\citeyear{gfpgan-webphoto}] and LFW-test~[\citeyear{lfw}]. 
RealPhoto60, WebPhoto, and LFW were tested at ×2 scale, and others at ×4.
}

\paragraph{Evaluation Metrics}
For the quantitative evaluation of super-resolution, we employed widely used perceptual quality metrics, including NIQE~[\citeyear{niqe}], CLIPIQA~[\citeyear{clipiqa}], MUSIQ~[\citeyear{musiq}], and QAlign~[\citeyear{qalign}], to compare the performance of the competing methods. 
Additionally, we provide the full-reference metrics including PSNR~[\citeyear{psnr}], SSIM~[\citeyear{ssim}], and LPIPS~[\citeyear{lpips}] in the full version.

\begin{figure*}[t]
    \centering
    \includegraphics[width=0.99\linewidth]{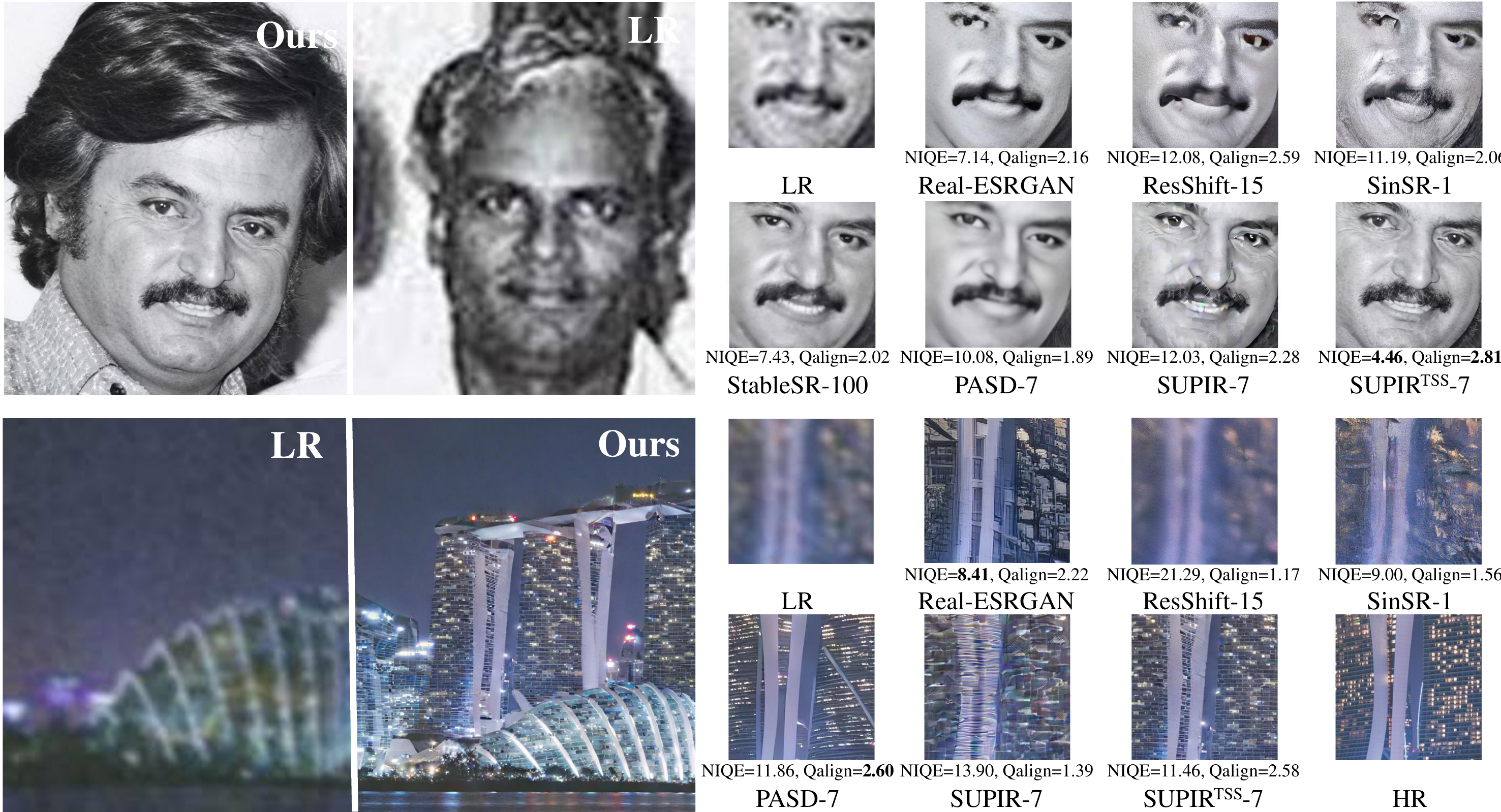}
    \caption{Qualitative comparison with state-of-the-art methods. Top: real-world sample from RealPhoto60 datasets. Bottom: synthetic sample from DIV2K valid datasets. More visual results are provided in the full version.}
    \label{fig:qualitative_comparison}
\end{figure*}

\subsection{Comparison with State-of-the-Art Methods}
For a comprehensive comparison, we integrate the TSS framework with three recent SOTA diffusion-based super-resolution (SR) methods: StableSR~[\citeyear{stablesr}], SUPIR~[\citeyear{yu2024supir}], and PASD~[\citeyear{yang2025pasd}]. Their performance is evaluated against state-of-the-art (SOTA) SR approaches, including Real-ESRGAN~[\citeyear{realesrgan}], BSRGAN~[\citeyear{bsrgan}], SwinIR~[\citeyear{swinir}], SinSR \noindent[\citeyear{wang2024sinsr}], and ResShift~[\citeyear{yue2024resshift}], using both synthetic and real-world datasets. Real-ESRGAN, BSRGAN, and SwinIR are traditional deep learning-based SR frameworks using CNNs or transformers, while SinSR and ResShift are diffusion-accelerated SR methods leveraging distillation and fine-tuning, respectively. For fairness, the results are obtained from official codebases and pre-trained models. Quantitative evaluation and qualitative comparisons are presented in Tab.~\ref{tab:sota_real} and Fig.~\ref{fig:qualitative_comparison}, respectively. As presented in Tab.~\ref{tab:sota_real}, our SUPIR$^{TSS}$ outperforms the state-of-the-art acceleration stra-
\begin{table}[H]
    
    \centering
    \resizebox{\linewidth}{!}{
    \setlength{\tabcolsep}{1.7mm}{
    \begin{tabular}{c|c|rrrrc}
        \toprule[1.5pt]
        Datasets & Methods & NIQE$\downarrow$ & CLIPIQA$\uparrow$ & MUSIQ$\uparrow$ & Qalign$\uparrow$ \\ 

        \midrule[1.5pt]
        \multirow{11}{*}{\makecell{DIV2K \\ SR (x4) \\ \quad}} & Real-ESRGAN & 4.87 & 0.5963 & 56.55 & 3.53 \\ 
        ~ & BSRGAN & 3.78 & 0.5804 & 60.25 & 3.68  \\ 
        ~ & SwinIR & \best{3.51} & 0.5677 & 58.27 & 3.92 \\ \cmidrule{2-6}
        ~ & ResShift (N=15)  & 7.68 & 0.5963 & 44.32 & 3.25 \\ 
        ~ & SinSR (N=1)  & 6.53 & \best{0.6745} & 55.28 & 3.66 \\ \cmidrule{2-6}
        ~ & StableSR(N=100)  & 7.94 & 0.3429 & 27.69 & 2.49 \\ 
        \rowcolor{gray!15} \cellcolor{white} ~ & StableSR$^{TSS}$(N=100)  & \textbf{7.16} & \textbf{0.3604} & \textbf{28.06} & \textbf{2.50} \\
        ~ & SUPIR(N=7)  & 5.02 & 0.3758 & 60.76 & 4.11 \\ 
        \rowcolor{gray!15} \cellcolor{white} ~ & SUPIR$^{TSS}$(N=7)  & \textbf{3.56} & \textbf{0.5244} & \best{\textbf{62.35}} & \best{\textbf{4.28}} \\ 
        ~ & PASD(N=7)  & 7.78 & 0.4071 & 39.79 & 3.03\\ 
        \rowcolor{gray!15} \cellcolor{white} ~ & PASD$^{TSS}$(N=7) & \textbf{6.11} & \textbf{0.4192} & \textbf{41.82} & \textbf{3.15} \\ 
        \midrule[1.5pt]
        
        \multirow{11}{*}{\makecell{RealSR \\ SR (x4) \\  \quad}} & Real-ESRGAN & 4.70 & 0.4818 & 59.50 & 3.92\\
        ~ & BSRGAN &  4.66 & 0.5399 & \best{63.37} & 3.86 \\
        ~ & SwinIR & 4.69 & 0.4636 & 59.40 & 3.85 \\\cmidrule{2-6}
        ~ & ResShift (N=15)& 7.43 & 0.5427 & 53.52 & 3.84 \\
        ~ & SinSR (N=1)& 6.24 & \best{0.6631} & 59.23 & 3.87 \\\cmidrule{2-6}
        ~ & StableSR(N=100) & 5.06 & 0.5530 & 60.99 & 3.90 \\
        \rowcolor{gray!15} \cellcolor{white} ~ & StableSR$^{TSS}$(N=100) & \textbf{4.98}  & 0.5691 & \textbf{61.15} & \textbf{3.93} \\
        ~ & SUPIR(N=7) & 6.44 & 0.4435 & 57.85 & 3.66 \\
        \rowcolor{gray!15} \cellcolor{white} ~ & SUPIR$^{TSS}$(N=7) & \textbf{4.76} & \textbf{0.5017} & \textbf{58.74} & \textbf{3.96}  \\
        ~ & PASD(N=7) & 4.99 & 0.5341 & 60.34 & 4.01 \\
        \rowcolor{gray!15} \cellcolor{white} ~ & PASD$^{TSS}$(N=7) & \best{\textbf{4.31}} & \textbf{0.5739} & \textbf{62.00} & \best{\textbf{4.08}} \\
        \midrule[1.5pt]

        \multirow{11}{*}{\makecell{DRealSR \\ SR (x4) \\  \quad}} & Real-ESRGAN & 4.35 & 0.5769 & 56.88 & \best{4.34} \\
        ~ & BSRGAN & 4.60 & 0.6125 & \best{58.55} & 4.31 \\
        ~ & SwinIR & 4.39 & 0.5668 & 56.93 & 4.33 \\\cmidrule{2-6}
        ~ & ResShift (N=15)& 6.03 & 0.6490 & 56.23 & 4.30 \\
        ~ & SinSR (N=1)& 5.51 & \best{0.7134} & 56.72 & 4.30 \\\cmidrule{2-6}
        ~ & StableSR(N=100) & 4.38 & 0.6472 & 55.88 & 4.26 \\
        \rowcolor{gray!15} \cellcolor{white} ~ & StableSR$^{TSS}$(N=100) & 4.73 & \textbf{0.6523} & \textbf{56.12} & \textbf{4.27} \\
        ~ & SUPIR(N=7) & 5.79 & 0.4760 & 52.69 & 4.09  \\
        \rowcolor{gray!15} \cellcolor{white} ~ & SUPIR$^{TSS}$(N=7) & \textbf{4.37} & \textbf{0.5362} & \textbf{54.26} & \textbf{4.27}  \\
        ~ & PASD(N=7) & 4.58 & 0.6612 & 58.21 & 4.30 \\
        \rowcolor{gray!15} \cellcolor{white} ~ & PASD$^{TSS}$(N=7) & \best{\textbf{3.95}} & \textbf{0.6923} & 57.58 & \textbf{4.32} \\

        \midrule[1.5pt]
        ~ & Real-ESRGAN & 3.92 & 0.5709 & 59.25 & 3.64 \\
        ~ & BSRGAN & 5.38 & 0.3305 & 45.46 & 2.11 \\
        \cmidrule{2-6}
        ~ & ResShift (N=15)& 6.59 & 0.6642 & 61.29 & 3.77 \\
        ~ & SinSR (N=1)& 5.91 & \best{0.7610} & 66.43 & 3.90 \\ \cmidrule{2-6}
        RealPhoto60 & StableSR(N=100) & 4.39 & 0.5484 & 55.41 & 3.65 \\
        \rowcolor{gray!15} \cellcolor{white} SR (x2) & StableSR$^{TSS}$(N=100) & \textbf{4.28} & \textbf{0.5575} & \textbf{56.40} & \textbf{3.68} \\
        ~ & SUPIR(N=7) & 5.80 & 0.4597 & 65.42 & 3.62 \\
        \rowcolor{gray!15} \cellcolor{white} ~ & SUPIR$^{TSS}$(N=7) & \best{\textbf{3.86}} & \textbf{0.6277} & \best{\textbf{67.86}} & \best{\textbf{4.38}} \\
        ~ & PASD(N=7) & 4.60 & 0.6244 & 63.85  & 3.99 \\
        \rowcolor{gray!15} \cellcolor{white} ~ & PASD$^{TSS}$(N=7) & \textbf{3.86} & \textbf{0.6427} & \textbf{66.38} & \textbf{4.22} \\

        \midrule[1.5pt]
        ~ & Real-ESRGAN & 5.92 & 0.4937 & 37.89 & 1.94 \\
        ~ & BSRGAN &  7.16 & 0.4316 & 38.86 & 1.78 \\
        \cmidrule{2-6}
        ~ & ResShift (N=15)& 10.24 & 0.4590 & 29.63 & 1.84 \\
        ~ & SinSR (N=1)& 7.94 & \best{0.6610} & 51.10 & 2.36 \\\cmidrule{2-6}
        WebPhoto & StableSR(N=100) & 7.54 & 0.3412 & 28.03 & 1.72 \\
        \rowcolor{gray!15} \cellcolor{white} SR (x2) & StableSR$^{TSS}$(N=100) & \textbf{7.00} & \textbf{0.3602} & \textbf{28.78} & \textbf{1.73} \\
        ~ & SUPIR(N=7) & 8.56 & 0.3935 & \best{60.42} & 2.95 \\
        \rowcolor{gray!15} \cellcolor{white} ~ & SUPIR$^{TSS}$(N=7) & \best{\textbf{5.19}} & \textbf{0.4815} & 58.72 & \best{\textbf{3.36}}  \\
        ~ & PASD(N=7) & 11.18 & 0.3172 & 28.73 & 1.85 \\
        \rowcolor{gray!15} \cellcolor{white} ~ & PASD$^{TSS}$(N=7) & \textbf{8.57} & \textbf{0.3485} & \textbf{31.25} & \textbf{2.06} \\
        \midrule[1.5pt]

        ~ & Real-ESRGAN & 5.06 & 0.5337 & 56.60 & 2.75 \\
        ~ & BSRGAN &  6.14 & 0.5566 & 58.48 & 2.60 \\
        \cmidrule{2-6}
        ~ & ResShift (N=15)& 8.42 & 0.5570 & 52.48 & 2.64 \\
        ~ & SinSR (N=1)& 6.97 & \best{0.7553} & 65.40 & 2.98 \\\cmidrule{2-6}
        LFW & StableSR(N=100) & 5.67 & 0.5246 & 55.51 & 3.18 \\
        \rowcolor{gray!15} \cellcolor{white} SR (x2) & StableSR$^{TSS}$(N=100) & \textbf{5.42} & \textbf{0.5416} & \textbf{56.50} & \textbf{3.22} \\
        ~ & SUPIR(N=7) & 6.21 & 0.4329 & 67.09 & 3.50 \\
        \rowcolor{gray!15} \cellcolor{white} ~ & SUPIR$^{TSS}$(N=7) & \best{\textbf{4.36}} & \textbf{0.5776} & \best{\textbf{67.23}} & \best{\textbf{4.22}}  \\
        ~ & PASD(N=7) & 6.51 & 0.5383 & 59.23 & 3.15 \\
        \rowcolor{gray!15} \cellcolor{white} ~ & PASD$^{TSS}$(N=7) & \textbf{5.10} & \textbf{0.5567} & \textbf{62.25} & \textbf{3.55} \\
        \bottomrule[1.5pt]
        
    \end{tabular}
    }
    }
    \caption{Quantitative comparison with SOTA real-world SR methods. `\textbf{Bold}' indicates improvement over the baseline, while `\underline{Underline}' denotes the best performance. SwinIR was evaluated only on x4 scale factors due to the lack of pretrained weights for x2.}
    \label{tab:sota_real}
\end{table}

\noindent tegies ReShift and SinSR on most benchmarks, achieving a 1.14 - 5.05 improvement in NIQE~[\citeyear{niqe}] and a 0.09 - 1.52 improvement in QAlign~[\citeyear{qalign}] across both synthetic and real datasets. 
Notably, while baseline SUPIR's 7-step QAlign results are initially inferior to ReShift and SinSR on real datasets, the incorporation of the TSS framework enables a QAlign improvement of 0.09 - 1.24 over SinSR and 0.12 - 1.58 over ReShift across all datasets. Although SinSR requires only a single iteration, its distillation strategy involves additional training costs, whereas TSS achieves comparable results without any extra training. Furthermore, compared to ReShift, which requires training and 15 inference steps, SUPIR$^{TSS}$ delivers superior performance across all metrics except CLIPIQA~[\citeyear{clipiqa}], achieving these results in just 7 iterations, less than half the steps.
Fig.~\ref{fig:qualitative_comparison} provides examples of synthetic and real degradation. Taking the first example in Fig.~\ref{fig:qualitative_comparison} as an example, most existing methods, including CNN- and Transformer-based approaches and diffusion acceleration strategies, 
fail to generate the region with high-frequency textures (wrinkles, beard, and eyebrow). In contrast, SUPIR$^{TSS}$ produces more realistic and richer high-frequency textures, as evidenced by both qualitative visual quality and quantitative results (Fig.~\ref{fig:qualitative_comparison}, Row. 2, Col. 4) In summary, both quantitative and qualitative analyses show that TSS significantly enhances the high-frequency restoration capabilities of recent diffusion-based SR methods, achieving state-of-the-art texture generation performance on both synthetic and real-world datasets without any additional training costs.

\begin{table}[t]
    \centering
    \resizebox{\linewidth}{!}{
    \setlength{\tabcolsep}{3.9mm}{
    \begin{tabular}{c|c|ccc}
        \toprule[1.2pt]
        \multirow{2}{*}{Strategy} & TDS & - & \checkmark & \checkmark \\
            ~ & SDS & - & - & \checkmark \\ 
        \midrule
        \multirow{4}{*}{Step=7} 
                 & NIQE $\downarrow$     & 5.80 & 3.99 & \textbf{3.86} \\ 
            ~ & CLIPIQA $\uparrow$    & 0.4597 & 0.6212 & \textbf{0.6277} \\ 
            ~ & MUSIQ $\uparrow$      & 65.42 & 67.53 & \textbf{67.86} \\
            ~ & Qalign $\uparrow$     & 3.62 & 4.35 & \textbf{4.38} \\ 
        \midrule

        \multirow{4}{*}{Step=10} & NIQE $\downarrow$ & 5.23 & 3.89 & \textbf{3.65} \\ 
            ~ &         CLIPIQA $\uparrow$& 0.5000 & 0.6391 & \textbf{0.6571} \\ 
            ~ &         MUSIQ $\uparrow$& 65.93 & 67.42 & \textbf{68.50} \\ 
            ~ &         Qalign $\uparrow$& 3.93 & 4.33 & \textbf{4.38} \\ 
        \midrule
            
        \multirow{4}{*}{Step=14} 
              & NIQE $\downarrow$        &  4.41 & 3.59 & \textbf{3.65} \\ 
            ~ & CLIPIQA $\uparrow$    &  0.5912 & 0.6438 & \textbf{0.6439} \\ 
            ~ & MUSIQ $\uparrow$      &  68.24 & 68.51 & \textbf{68.61} \\ 
            ~ & Qalign $\uparrow$     &  4.26 & 4.31 & \textbf{4.33} \\ 
        \bottomrule[1.2pt]
    \end{tabular}
    }
    }
    \caption{Comparison of SUPIR with different sampling strategies. }
    \label{tab:ablation}
\end{table}

\begin{figure}[t]
    \centering
    \includegraphics[width=\linewidth]{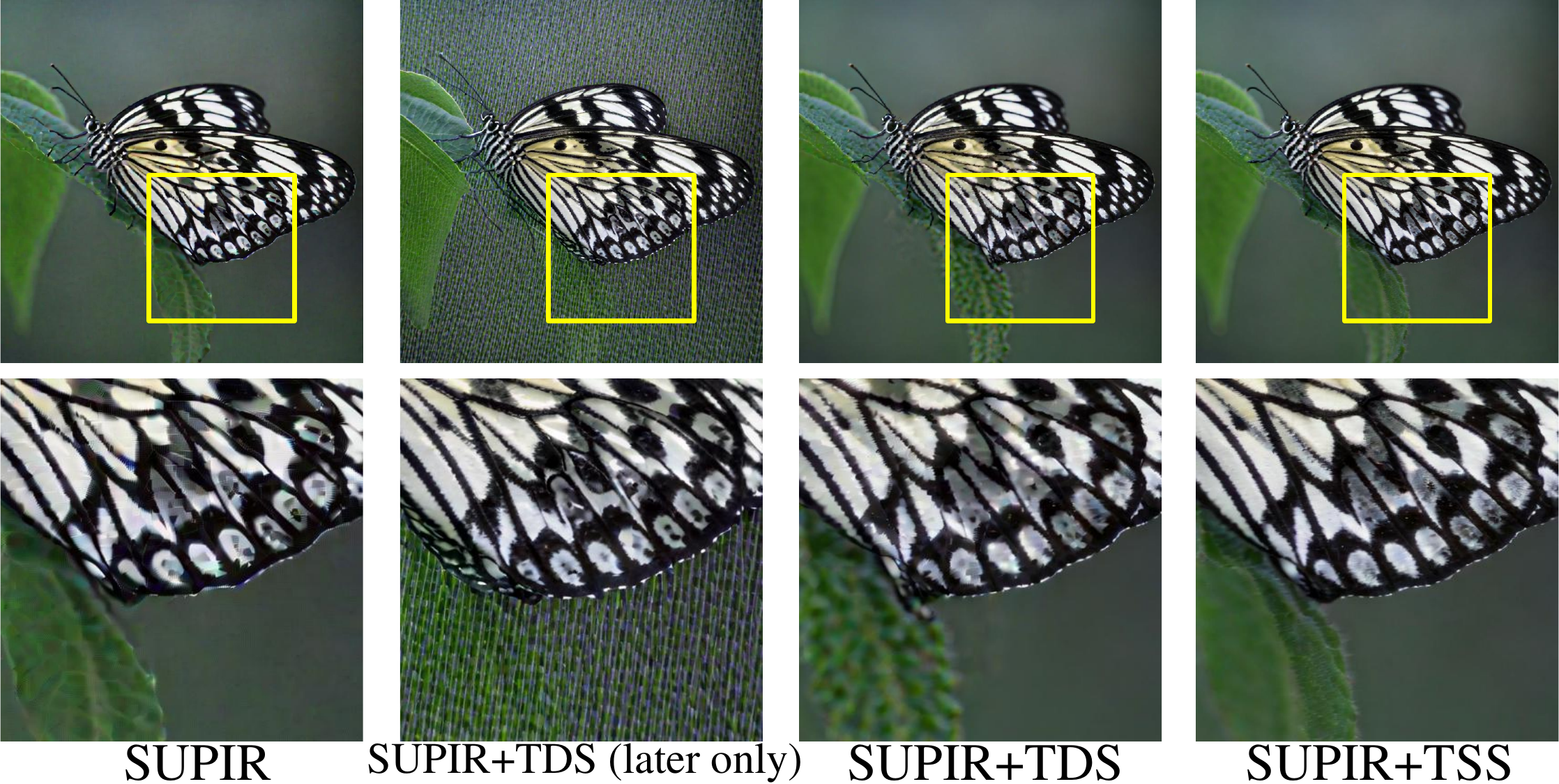}
    \caption{Visual comparison of SUPIR-7 with different sampling strategies on real samples in RealPhoto60.}
    \label{fig:ablation}
\end{figure}

\subsection{Ablation Study}
\begin{figure}[t]
    \centering
    \includegraphics[width=\linewidth]{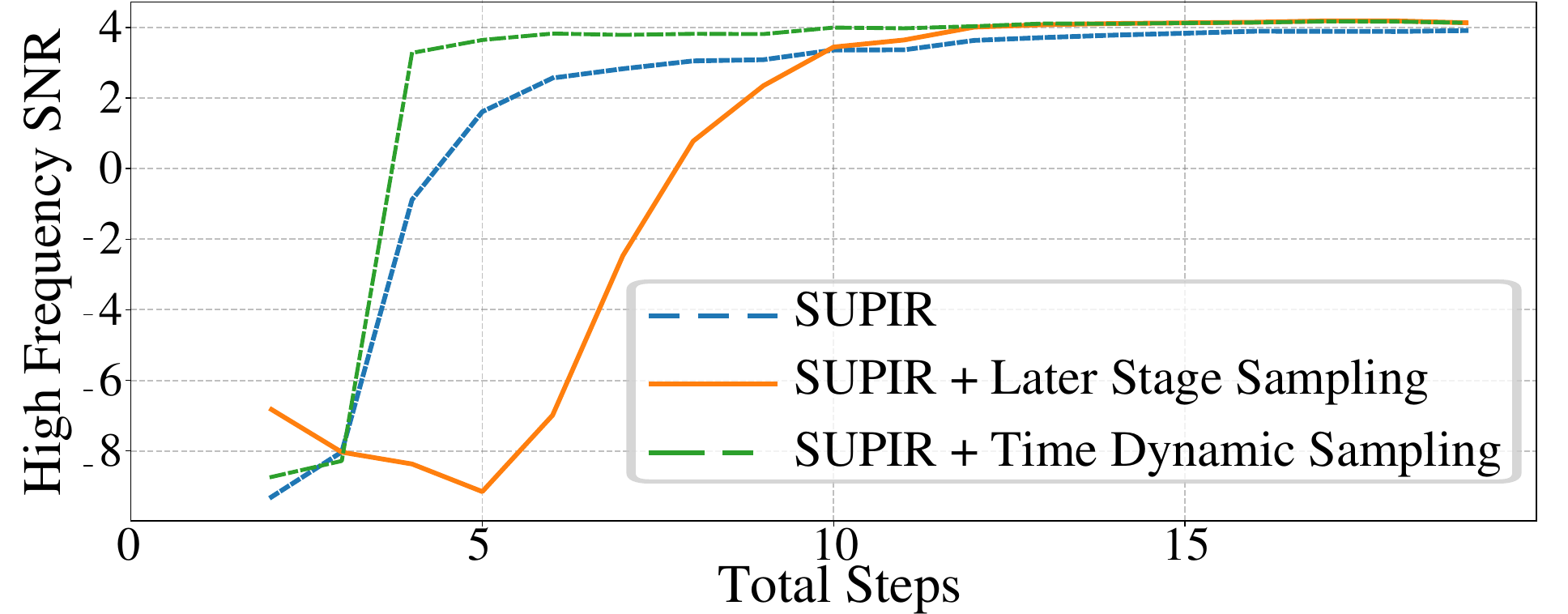}
    \caption{Comparison of SNR for high-frequency components at different steps counts when using different sampling strategies.}
    \label{fig:hfs_cmp}
\end{figure}

\paragraph{Effectiveness of Time Dynamic Sampling}
To validate the effectiveness of Time Dynamic Sampling (TDS), we compare SUPIR with its TDS-enhanced version on the RealPhoto60 dataset. As shown in Tab.~\ref{tab:ablation}, TDS improves quantitative metrics, achieving a 20\% increase in QAlign[\citeyear{qalign}] and a 35\% increase in CLIPIQA~[\citeyear{clipiqa}] at 7 steps.
Fig.~\ref{fig:ablation} (Col. 3 vs. 1) visually compares SUPIR results with and without TDS on real samples. TDS enhances the butterfly texture reconstruction while uniform sampling introduces artifacts like color blocking.
For further analysis, Fig.~\ref{fig:hfs_cmp} illustrates the relationship between high-frequency SNR and the number of denoising steps. Compared to the original SUPIR (blue), SUPIR+TDS (green) consistently achieves higher SNR at the same step count, even with fewer total steps, indicating its effectiveness in high-frequency signal restoration.

\paragraph{Necessity of Early Stage Sampling}
To verify the need for early-stage sampling, we compare SUPIR with late-stage-only sampling (\(a = T\) in Eq.~\ref{eq:non-uniform_sampling}) and Time Dynamic Sampling (TDS) on the RealPhoto60. As shown in Tab.~\ref{tab:early_stage}, late-stage-only sampling fails to improve NIQE and results in a 0.3 drop in QAlign, while TDS nearly doubles the MUSIQ gain by balancing early and late sampling.  
Fig.~\ref{fig:hfs_cmp} compares the high-frequency SNR of both methods. Due to insufficient early-stage denoising, late-stage-only sampling retains excessive HF noise at lower steps, sometimes underperforming the baseline. In contrast, TDS, with concentrated sampling in both early and late stages, effectively boosts HF SNR with fewer steps.  
Fig.~\ref{fig:ablation} (Col. 2 vs. 3) further illustrates this effect. While late-stage-only sampling enhances local texture, fewer steps in the early stage leave residual noise, causing visible artifacts. TDS, by ensuring adequate sampling across both stages, enables realistic texture generation without artifacts.

\begin{table}[t]
    \centering
    \resizebox{\linewidth}{!}{
    \setlength{\tabcolsep}{2.7mm}{
    \begin{tabular}{c|ccc}
        \toprule[1.2pt]
            Metrics & Uniform & Later Stage Only & TDS \\ 
            \midrule
            NIQE $\downarrow$     & 5.80 & 5.80 (-0.00) & \textbf{3.99} (\textcolor{red}{-1.81}) \\ 
            CLIPIQA $\uparrow$    & 0.4597 & 0.6032 (\textcolor{red}{+0.1435}) & \textbf{0.6212} (\textcolor{red}{+0.1615}) \\ 
            MUSIQ $\uparrow$      & 65.42 & 66.58 (\textcolor{red}{+1.16}) & \textbf{67.53} (\textcolor{red}{+2.11}) \\
            Qalign $\uparrow$     & 3.62 & 3.32 (\textcolor{blue}{-0.30}) & \textbf{4.35} (\textcolor{red}{+0.73}) \\ 
            \bottomrule[1.2pt]
    \end{tabular}
    }
    }
    \caption{Ablation study on the necessity of early-stage sampling. \textcolor{red}{Red} for improvement, and \textcolor{blue}{blue} for degradation.}
    \label{tab:early_stage}
    
\end{table}

\begin{table}[b]
    \centering
    \resizebox{\linewidth}{!}{
    \setlength{\tabcolsep}{1.9mm}{
    \begin{tabular}{c|cccc}
        \toprule[1.2pt]

            \multirow{2}{*}{Metrics} & \multirow{2}{*}{Uniform}  & \multicolumn{3}{c}{Non-Uniform} \\  \cmidrule{3-5}
            ~ & ~ & Trigonometric & Exponential & Polynomial \\ \midrule
            NIQE $\downarrow$    & 5.80 & 3.96 & 4.36 & \textbf{3.99} \\ 
            CLIPIQA $\uparrow$    & 0.4597 & \textbf{0.6224} & 0.6033 & 0.6212 \\ 
            MUSIQ $\uparrow$      & 65.42 & 67.42 & 66.11 & \textbf{67.53} \\ 
            Qalign $\uparrow$     & 3.62 & 4.33 & 4.25 & \textbf{4.35} \\  
            \bottomrule[1.2pt]
    \end{tabular}
    }
    }
    \caption{Comparison of non-uniform functions used in TDS. The detailed function description is in the full version.}
    \label{tab:non-uniform_function}
\end{table}

\begin{figure}[t]
    \centering
    \includegraphics[width=\linewidth]{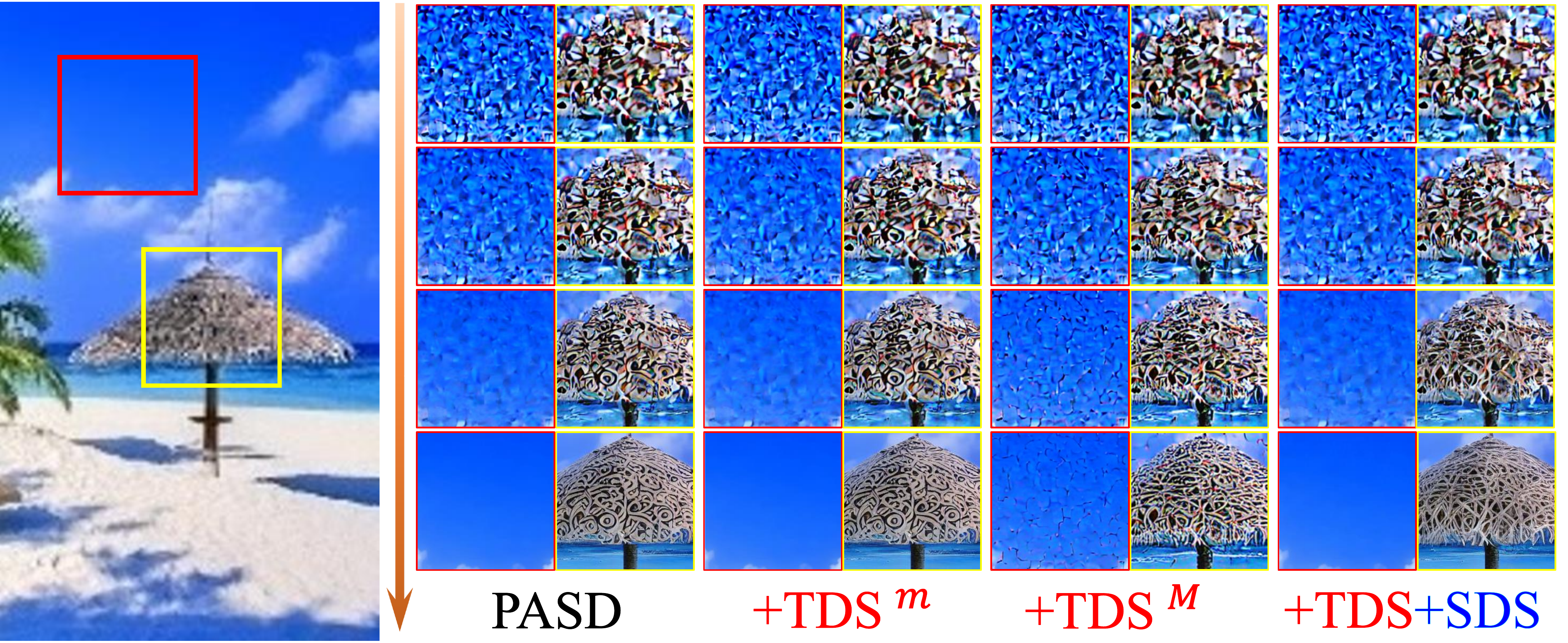}
    \caption{\modified{Last 4 steps in PASD's 7-step denoising process. "+TDS" strategies fixes param $a$ and $n$ to their \textbf{m}in and \textbf{M}ax value.}}
    \label{fig:pasd_denoising_log}
\end{figure}

\paragraph{Non-Uniform Function Selection}
We evaluated the impact of different non-uniform functions on sampling performance, including trigonometric, exponential, and polynomial functions. As shown in Tab.~\ref{tab:non-uniform_function}, all three functions outperform the baseline with a 0.6 - 0.75 QAlign improvement, showing that non-uniform sampling aligns with high-frequency signal recovery regardless of the function type. For better controllability, we select the polynomial function as the final setting.

\paragraph{Effectiveness of Spatial Dynamic Sampling}
To assess the necessity of Spatial Dynamic Sampling (SDS), we compare SUPIR with TDS alone and the full TSS (TDS+SDS) on RealPhoto60. As shown in Tab.~\ref{tab:ablation} (Col. 2 and 3), the inclusion of SDS brings quantitative metric improvements over TDS alone.
Moreover, Fig.~\ref{fig:ablation} further illustrates this effect. TDS alone (Col. 3) applies a spatial-unified scheduler, resulting in less sharp butterfly wings and an insufficiently smooth background. In contrast, SDS enables spatial-wise adaptive sampling, enhancing both texture-rich and smooth regions.

\paragraph{SDS in the Denoising Process} For a more intuitive explanation, we recorded the denoising process of PASD with different sampling strategies in Fig.~\ref{fig:pasd_denoising_log}. Excessive inhomogeneity (Col. 3) adds noise to smooth regions, while insufficient inhomogeneity (Cols. 1, 2) fails to generate textures. In contrast, SDS (Col. 4) adaptively balances inhomogeneity, ensuring optimal reconstruction of both smooth and textured areas.

%% file: data/conclusion.tex

\section{Conclusion}
In this work, we explore key insights in the denoising process: high-frequency components require focused optimization in early and late iterations, while spatially varying content necessitates adaptive strategies for effective restoration. Based on these findings, we propose Time-Spatial-aware Sampling (TSS), a training-free, content-adaptive sampling strategy to accelerate diffusion-based image super-resolution. By leveraging temporal and spatial dependencies in high-frequency recovery, TSS enhances texture restoration while significantly reducing computational costs. Compatible with various diffusion SR frameworks, it achieves state-of-the-art results with half the steps of recent accelerated methods. 

%% file: data/appendix.tex
\setcounter{table}{4}
\setcounter{figure}{8}
\section{Appendix}
\subsection{Details of Non-Uniform Function in Ablation Study}
\label{sec:sample_function}
In the ablation study, we compare the performance of Time Dynamic Sampling (TDS) using the Polynomial function with other kinds of nonlinear functions including trigonometric and exponential functions. The detailed form of the trigonometric function $f_{Tri}$ and exponential function $f_{Exp}$ are as follows:
\begin{equation}
    f_{Tri}(t) = \begin{cases}
    -a\cos(\frac{\pi t}{2a})+a, t < a\\
    -a\sin(\frac{\pi(t-a)}{2(T-a)})+T, t \geq a\\
    \end{cases}
\end{equation}
\begin{equation}
    f_{Exp}(t) =\frac{T}{1 + \exp(-k(t - \frac{T}{2}))},
\end{equation}
where $a$ is set as the same in TDS and $k=0.004$. 
\modified{The metric results for the three functions are shown in Tab.~\ref{tab:non-uniform_function}. Beyond performance metrics, controllability is also considered. The trigonometric function allows control over the early and late sampling ranges but cannot approximate uniform sampling by adjusting hyperparameters. In contrast, the exponential function can approximate uniform sampling but lacks control over the early and late sampling ranges. The polynomial function, however, offers both capabilities, making it the most controllable option. Therefore, considering controllability and performance, we select the polynomial function as the final choice. It is worth noting that, the choice of polynomial TDS functions and linear variance mapping in SDS is driven primarily by strategic design considerations, rather than dependence on specific functional forms. The non-uniform function is not restricted to a particular formulation—various alternatives can yield effective improvement as long as they satisfy the distributional conditions of pre- and post-centralized sampling. The current setup was chosen for its effectiveness, controllability, and simplicity, and we encourage future exploration of more advanced sampling and mapping strategies.}

\subsection{Detailed Implementation Settings and Hyperparameters}
Here, we provide the detailed hyperparameters and settings for the experiments in the main paper. To ensure fairness and generalizability, as shown in Tab.~\ref{tab:sota_real}, we evaluate our TSS strategy on three different anchor methods. Since each anchor method contains distinct backbones and base samplers, small differences in the values of the hyperparameters are taken to achieve the best performance.
\begin{table}[t]
    \centering
    \resizebox{\linewidth}{!}{
    \setlength{\tabcolsep}{2mm}{
    \begin{tabular}{c|ccc}
    \toprule[1.2pt]
        Method & StableSR~[\citeyear{stablesr}] & PASD~[\citeyear{yang2025pasd}] & SUPIR~[\citeyear{yu2024supir}] \\ \midrule
        Backbone & \makecell{Stable Diffusion \\2.1~[\citeyear{rombach2022sd}]} & \makecell{Stable Diffusion \\1.5~[\citeyear{rombach2022sd}]} & \makecell{Stable Diffusion \\XL~[\citeyear{podell2023sdxl}]} \\ \midrule
        Sampler & DDPM+~[\citeyear{ho2020ddpm}] & UniPC~[\citeyear{zhao2024unipc}] & EDM Euler~[\citeyear{edm}]\\ \midrule
        $n_\text{min},n_\text{max}$ & 1,1.2 & 1,2 & 2.2,2.5 \\ 
        $a_\text{min},n_\text{max}$ & 0.45,0.65 & 0.4,0.6 & 0.58,0.63 \\ 
        \bottomrule[1.2pt]
    \end{tabular}
    }
    }
    \caption{Hyperparameters and settings for the experiments.}
\end{table}

\begin{table}[t]
    \centering
    \resizebox{\linewidth}{!}{
    \setlength{\tabcolsep}{1mm}{
    \begin{tabular}{c|c|c|rrrr}
        \toprule[1.2pt]
        \makecell{Anchor\\Method} & Step & Sampling & NIQE $\downarrow$ & CLIPIQA $\uparrow$ & MUSIQ $\uparrow$ & Qalign $\uparrow$\\ \midrule
        \multirow{4}{*}{StableSR} & \multirow{2}{*}{7} & DDIM & 7.12 & 0.6471 & 26.69 & 1.60 \\ 
        ~ & ~ & DDIM+TSS & \cellcolor{gray!15}\textbf{5.70} & \cellcolor{gray!15}\textbf{0.6970} & \cellcolor{gray!15}\textbf{37.91} & \cellcolor{gray!15}\textbf{1.67} \\ \cmidrule{2-7}
        ~ & \multirow{2}{*}{15} & DDIM & 5.43 & 0.6976 & 39.81 & 1.86 \\ 
        ~ & ~ & DDIM+TSS & \cellcolor{gray!15}\textbf{4.13} & \cellcolor{gray!15}\textbf{0.7141} & \cellcolor{gray!15}\textbf{52.33} & \cellcolor{gray!15}\textbf{1.89} \\ \midrule
        
        \multirow{4}{*}{PASD} & \multirow{2}{*}{7}& UniPC & 11.18 & 0.3172 & 28.73 & 1.85 \\ 
        ~  & ~ & UniPC+TSS & \cellcolor{gray!15}\textbf{8.57} & \cellcolor{gray!15}\textbf{0.3485} & \cellcolor{gray!15}\textbf{31.26} & \cellcolor{gray!15}\textbf{2.07} \\ \cmidrule{2-7}
        ~ & \multirow{2}{*}{15} & UniPC & 10.76 & 0.3188 & 28.73 & 1.88 \\ 
        ~  & ~ & UniPC+TSS & \cellcolor{gray!15}\textbf{7.08} & \cellcolor{gray!15}\textbf{0.3694} & \cellcolor{gray!15}\textbf{31.41} & \cellcolor{gray!15}\textbf{2.11} \\ 
        \bottomrule[1.2pt]
    \end{tabular}
    }
    }
    \caption{Comparison of generative acceleration sampler with and without our Time-Spatial-aware Sampling strategy.}
    \label{tab:generative}
\end{table}
\subsection{Relation with General Accelerated Sampling Strategies}
To investigate the relationship between our strategy and commonly used generative acceleration techniques, we apply both generative acceleration strategies (DDIM~[\citeyear{song2020ddim}] and UniPC~[\citeyear{zhao2024unipc}]) with our TSS to StableSR~[\citeyear{stablesr}] and PASD~[\citeyear{yang2025pasd}] on WebPhoto~[\citeyear{gfpgan-webphoto}] datasets. The results of DDIM and UniPC sampler acceleration are generated by the implementation of the official repository. As shown in Tab.~\ref{tab:generative}, regardless of whether the classic DDIM or the widely recognized UniPC sampler is used, integrating them with TSS brings consistent improvement on all evaluation metrics. This shows that our method is compatible with existing accelerated samplers and provides complementary benefits. While generative acceleration enhances efficiency based on model assumptions, TSS leverages task-specific priors to further improve the performance of diffusion-based SR, making them mutually beneficial when combined.

\begin{table}[b]
\small
\centering

\resizebox{\linewidth}{!}{
\setlength{\tabcolsep}{7.9mm}{
    \begin{tabular}{c|cc}
    \toprule[1.2pt]
    Method & DIV2K & RealSR \\
    \midrule
    ResShift$_{N=15}$ & 105.65 & 61.53 \\
    SinSR$_{N=1}$ & 101.27 & 54.89 \\
    \midrule
    PASD$_{N=7}$ & 98.06 & 54.06 \\
    PASD$^{TSS}_{N=7}$ & \cellcolor{gray!15}\textbf{92.70} & \cellcolor{gray!15}\underline{\textbf{50.91}} \\
    SUPIR$_{N=7}$ & 87.67 & 70.32 \\
    SUPIR$^{TSS}_{N=7}$ & \cellcolor{gray!15}\underline{\textbf{81.08}} & \cellcolor{gray!15}\textbf{52.99} \\
    \bottomrule[1.2pt]
    \end{tabular}
    }
}
\caption{FID ($\downarrow$) results comparison of different methods on DIV2K and RealSR datasets.}
\end{table}
\subsection{\modified{FID Comparisons}}
As FID requires enough samples for reliable evaluation, and image super-resolution benchmarks typically contain around 100 images, we follow standard protocols from prior work in the paper and calculate FID ($\downarrow$) results are shown below. TSS achieves clear improvements over baseline methods.

\subsection{\modified{Comparison with Other Acceleration Method Adjusting Denoise Schedule}}

 \modified{To further evaluate the effectiveness of our TSS strategy, we conduct a comparison with DM-NonUniform~\cite{dmununiform}, which modifies the denoising schedule by formulating an optimization problem to determine optimal time steps tailored to a specific numerical ODE solver. specifically, we apply DM-NonUniform and TSS to the SUPIR method and evaluate their performance on the DIV2K dataset. As shown in Tab.~\ref{tab:quantitative_accelaration}, TSS consistently outperforms DM-NonUniform in terms of perceptual metrics, thereby further validating the efficacy of our proposed approach.}

\begin{table}[h]
\small
\centering
\resizebox{\linewidth}{!}{
\begin{tabular}{c|cccc}
\toprule[1.2pt]
~ & NIQE$\downarrow$ & MUSIQ$\uparrow$ & CLIPIQA$\uparrow$ & Qalign$\uparrow$  \\
\midrule
SUPIR$^{DM}_{N=7}$& 4.64 & 61.16 & \textbf{0.5735} & 3.41 \\
\rowcolor{gray!15}
SUPIR$^{TSS}_{N=7}$ (Ours) & \textbf{3.56} & \textbf{62.35} & 0.5244 & \textbf{4.28} \\
\bottomrule[1.2pt]
\end{tabular}
}
\caption{Comparison of TSS with DM-NonUniform using SUPIR on DIV2K dataset.}
\label{tab:quantitative_accelaration}
\end{table}

\subsection{\modified{Quantitative Efficiency Improvement Results}}

\modified{To intuitively assess the efficiency gains of TSS in acceleration, we compare PASD$^{TSS}_{N=7}$ with the default PASD$_{N=20}$ on the RealPhoto60 dataset using an NVIDIA RTX 3090 GPU. Given that the computational cost is approximately proportional to the number of denoising steps, TSS achieves superior performance with only 7 steps—representing a reduction of over 65\%—thus demonstrating its effectiveness in accelerating inference without compromising quality. Furthermore, as presented in Tab.~\ref{tab:quantitative_accelaration}, the quantitative results for SUPIR under the baseline setting (SUPIR$_{N=14}$, 1st column, 3rd row) and the TSS-accelerated version (SUPIR$^{TSS}_{N=7}$, 3rd column, 1st row) exhibit a similar trend to PASD. These findings further confirm the effectiveness of TSS in enhancing the performance of various diffusion-based super-resolution methods.}

\begin{table}[t]
\small
\centering
\resizebox{\linewidth}{!}{
\begin{tabular}{c|c|c}
\toprule[1.2pt]
Metric & PASD$_{N=20}$ & \cellcolor{gray!15}PASD$^{TSS}_{N=7}$ \\
\midrule
NIQE $\downarrow$ & 4.31 & \cellcolor{gray!15}\textbf{3.87} \\
MUSIQ $\uparrow$ & 63.22 & \cellcolor{gray!15}\textbf{66.38} \\
CLIPIQA $\uparrow$ & 0.6089 & \cellcolor{gray!15}\textbf{0.6427} \\
Qalign $\uparrow$ & 4.09 & \cellcolor{gray!15}\textbf{4.22} \\
\midrule
Avg Time Per Image(s) $\downarrow$ & 10.08 & \cellcolor{gray!15}\textbf{4.61} (2.2$\times$ speed up) \\
\bottomrule[1.2pt]
\end{tabular}
}
\caption{Comparison of PASD and PASD$^{TSS}$ in terms of quality metrics and efficiency.}
\label{tab:quantitative_accelaration}
\end{table}

\subsection{Robustness of TDS to Hyperparameters}
\modified{We evaluate the robustness of TDS to hyperparameters (HP), including the number of steps $n$, the sampling range $a$, by testing different settings using PASD on RealSR dataset. specifically, we set $n$ to 1.60, 1.20, and 1.80, and $a$ to 0.30, 0.50, and 0.70, while keeping the other parameters unchanged. As shown in Tab.~\ref{tab:hyperparameters}, despite some performance variation, most "+TDS" settings consistently surpass the baseline, indicating robust improvement by TDS and its potential for adjustment. This characteristic of TDS offers a solid foundation and motivation for the subsequent integration of SDS.}

\begin{table}[h]
\small
\centering
\resizebox{\linewidth}{!}{
\setlength{\tabcolsep}{1mm}{
\begin{tabular}{c|c|ccccc}
    \toprule[1.2pt]
    \multirow{3}{*}{Settings} & \multirow{3}{*}{\makecell{w/o TDS \\ (baseline)}} & \multicolumn{5}{c}{+TDS} \\ \cline{3-7}
    ~ & ~ & n=1.60, & n=1.60, & n=1.60, & n=1.20, & n=1.80,\\ 
    ~ & ~ & a=0.30 & a=0.70 & a=0.50 & a=0.50 & a=0.50 \\
    \midrule
    NIQE $\downarrow$    & 4.99      & 4.14      & 4.69      & 4.12      & 4.63      & 4.42 \\
    MUSIQ   $\uparrow$& 60.34     & 61.80     & 59.88     & 60.81     & 61.47     & 59.68 \\
    CLIPIQA $\uparrow$ & 0.5341    & 0.5778    & 0.5772    & 0.5713    & 0.5576    & 0.5472 \\
    Qalign  $\uparrow$& 4.01      & 4.00      & 3.94      & 3.98      & 4.06      & 3.86 \\
    \bottomrule[1.2pt]
\end{tabular}
}
}
\caption{PASD+TDS results under varying hyperparameters.}
\label{tab:hyperparameters}
\end{table}

\subsection{More Comparisons with State-of-the-Art Methods}
\begin{figure*}
    \centering
    \includegraphics[width=\linewidth]{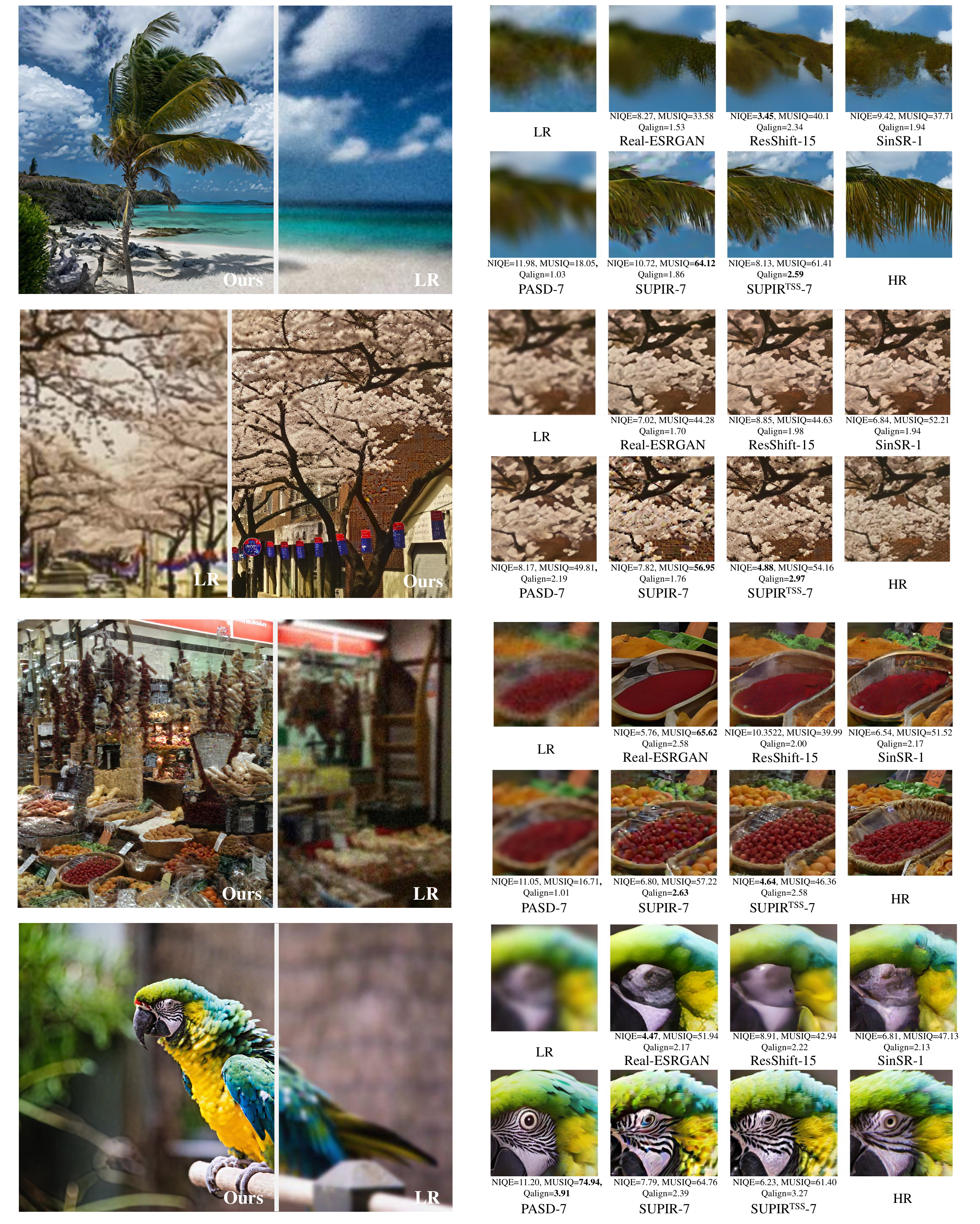}
    \caption{More visual comparison with state-of-the-art Real-world SR methods.}
\label{fig:extra_compare_1}
\end{figure*}

\begin{figure*}
    \centering
    \includegraphics[width=\linewidth]{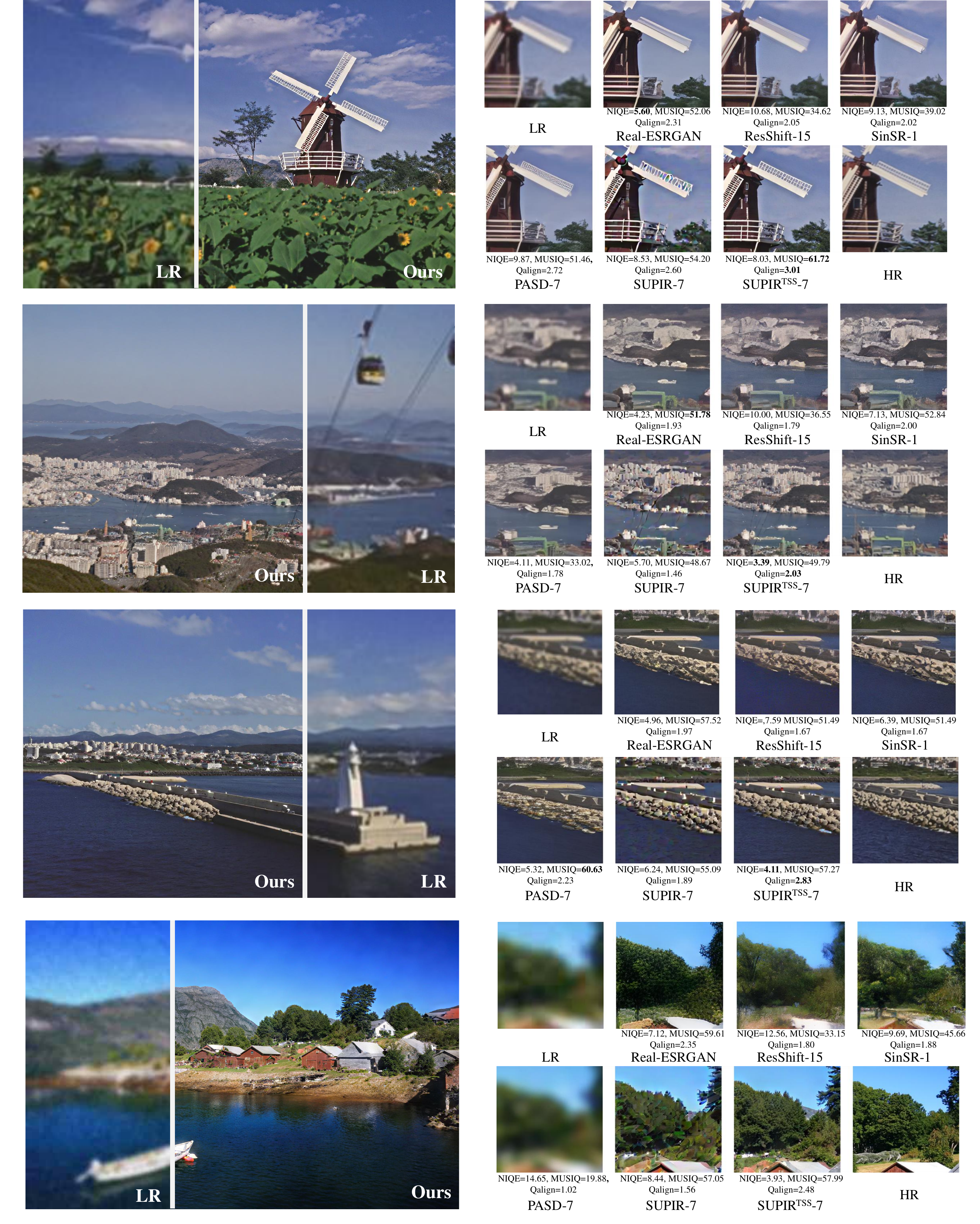}
    \caption{More visual comparison with state-of-the-art Real-world SR methods.}
\label{fig:extra_compare_2}
\end{figure*}

\begin{figure*}
    \centering
    \includegraphics[width=\linewidth]{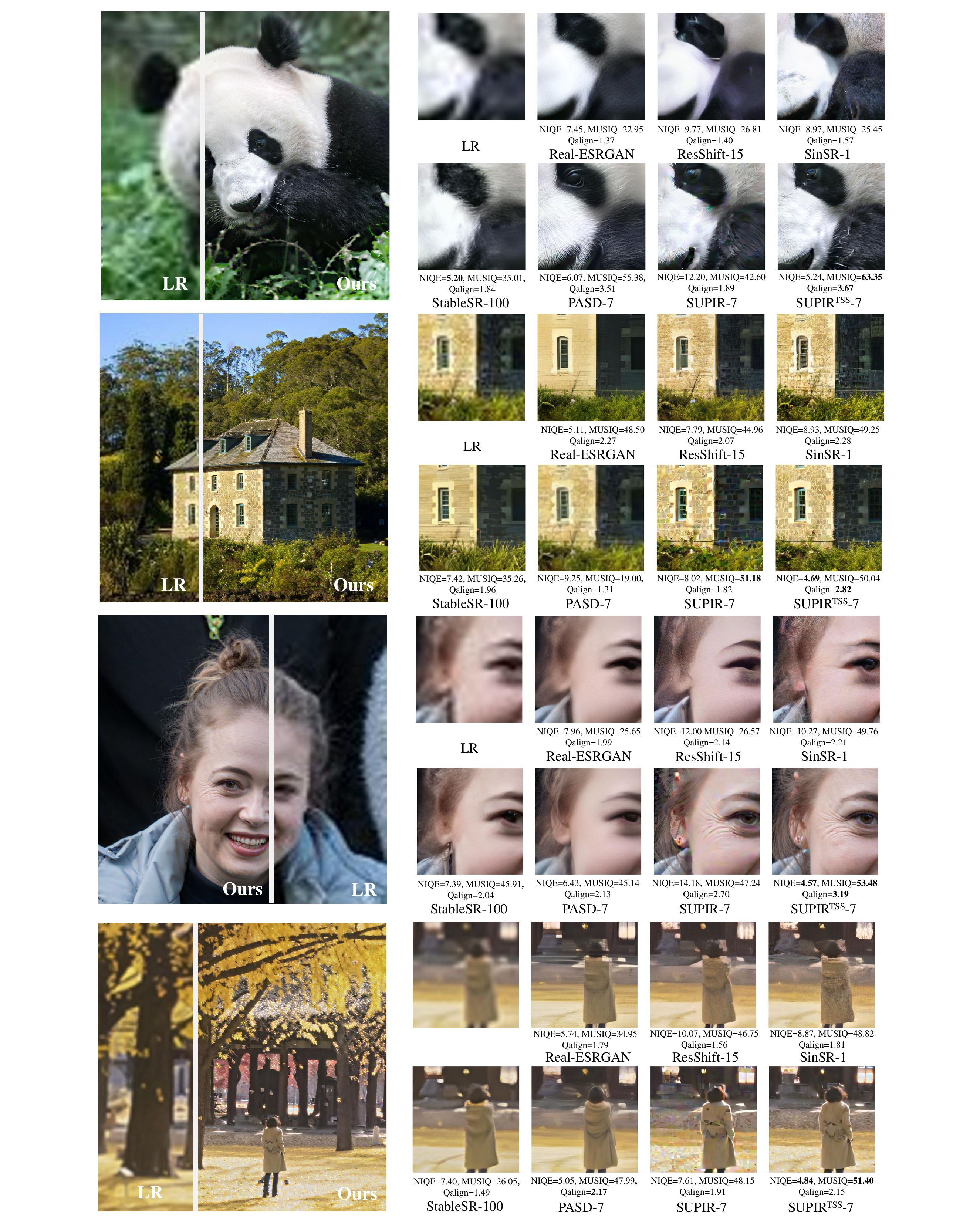}
    \caption{More visual comparison with state-of-the-art Real-world SR methods.}
\label{fig:extra_compare_3}
\end{figure*}

In this section, we provide more comparisons with state-of-the-art methods, including quantitative and qualitative results in Tab.~\ref{tab:sota_real_extra} and Fig.~\ref{fig:extra_compare_1}, Fig.~\ref{fig:extra_compare_2}, and Fig.~\ref{fig:extra_compare_3}.

\subsection{Limitation Discussion and Future Work}
\modified{First, as TDS is formulated as a modification to the denoising scheduler, it is not directly applicable to recent one-step methods such as SinSR~\cite{wang2024sinsr}. In contrast, the spatially diverse concept of SDS offers greater flexibility and may be more readily extended to such frameworks. Moreover, while the hyperparameters of the TSS strategy show robustness across various test datasets, they exhibit sensitivity to model architecture, as indicated in Tab.~\ref{tab:hyperparameters}. Regarding spatial dynamic embedding, we find that when the timestep interval between neighboring becomes excessively large, it can negatively impact generation quality. This observation implies that although inference-time adaptation is compatible with existing pre-trained models, certain boundary limitations remain due to the lack of training-phase adaptation. Consequently, learning spatially dynamic time steps during training presents a reasonable direction for future work. In addition, as discussed in Sec~\ref{sec:sample_function}, the design of both the sampling function and spatial dynamics in this study is primarily guided by empirical insights and application simplicity. A more rigorous and theoretically grounded formulation of these components would be a valuable avenue for further research. Finally, the TSS strategy is not restricted to super-resolution tasks; it holds potential for broader diffusion-based applications such as image inpainting, text-to-image generation, and video processing, which we plan to investigate in future work.}

\begin{table*}
    
    \centering
    \resizebox{\linewidth}{!}{
    \setlength{\tabcolsep}{5mm}{
    \begin{tabular}{c|c|rrrr>{\color{gray}}r>{\color{gray}}r>{\color{gray}}r}
        \toprule[1.2pt]
        Datasets & Methods & NIQE$\downarrow$ & CLIPIQA$\uparrow$ & MUSIQ$\uparrow$ & Qalign$\uparrow$& PSNR$\uparrow$& SSIM$\uparrow$& LPIPS$\downarrow$ \\ \midrule[1.2pt]
        \multirow{11}{*}{\makecell{DIV2K \\ SR (x4) \\ \quad}} & Real-ESRGAN & 4.87 & 0.5963 & 56.55 & 3.53 & 20.79 & 0.5425 & 0.4739 \\ 
        ~ & BSRGAN & 3.78 & 0.5804 & 60.25 & 3.68 & \best{22.04} & 0.5615 & 0.4253 \\ 
        ~ & SwinIR & 3.51 & 0.5677 & 58.27 & 3.92 & 21.25 & 0.5604 & \best{0.4181} \\ \cmidrule{2-9}
        ~ & ResShift (N=15)  & 7.68 & 0.5963 & 44.32 & 3.25 & 20.74 & 0.4891 & 0.5976 \\ 
        ~ & SinSR (N=1) & 6.53 & \best{0.6745} & 55.28 & 3.66 & 20.32 & 0.4635 & 0.5379 \\ \cmidrule{2-9}
        ~ & StableSR(N=100)  & 7.94 & 0.3429 & 27.69 & 2.49 & 21.39 & 0.5676 & 0.5732 \\ 
        \rowcolor{gray!15} \cellcolor{white} ~ & StableSR$^{TSS}$(N=100)  & \textbf{7.16} & \textbf{0.3604} & \textbf{28.06} & \textbf{2.50} & 21.34 & 0.5634 & \textbf{0.5695}\\
        ~ & SUPIR(N=7)  & 5.02 & 0.3758 & 60.76 & 4.11 & 19.86 & 0.4723 & 0.4909\\ 
        \rowcolor{gray!15} \cellcolor{white} ~ & SUPIR$^{TSS}$(N=7)  & \best{\textbf{3.56}} & \textbf{0.5244} & \best{\textbf{62.35}} & \best{\textbf{4.28}} & \textbf{20.34} & \textbf{0.4955} & \textbf{0.4354}\\ 
        ~ & PASD(N=7)  & 7.78 & 0.4071 & 39.79 & 3.03 & 21.28 & \best{0.5684} & 0.5442 \\ 
        \rowcolor{gray!15} \cellcolor{white} ~ & PASD$^{TSS}$(N=7) & \textbf{6.11} & \textbf{0.4192} & \textbf{41.82} & \textbf{3.15} & 20.98 & 0.5544 & \textbf{0.5155} \\ \midrule[1.2pt]
        
        \multirow{11}{*}{\makecell{RealSR \\ SR (x4) \\  \quad}} & Real-ESRGAN & 4.70 & 0.4818 & 59.50 & 3.92 & 24.71 & 0.7512 & 0.2698\\
        ~ & BSRGAN &  4.66 & 0.5399 & \best{63.37} & 3.86 & \best{25.38} & 0.7524 & 0.2675\\
        ~ & SwinIR & 4.69 & 0.4636 & 59.40 & 3.85 & 24.97 & \best{0.7574} & 0.2586 \\\cmidrule{2-9}
        ~ & ResShift (N=15)& 7.43 & 0.5427 & 53.52 & 3.84 & 24.18 & 0.6727 & 0.4576 \\
        ~ & SinSR (N=1)& 6.24 & \best{0.6631} & 59.23 & 3.87 & 24.19 & 0.6769 & 0.4013 \\\cmidrule{2-9}
        ~ & StableSR(N=100) & 5.06 & 0.5530 & 60.99 & 3.90 & 24.30 & 0.7401 & 0.2576 \\
        \rowcolor{gray!15} \cellcolor{white} ~ & StableSR$^{TSS}$(N=100) & \textbf{4.98}  & 0.5691 & \textbf{61.15} & \textbf{3.93} & 24.17 & 0.7338 & \textbf{\best{0.2525}} \\
        ~ & SUPIR(N=7) & 6.44 & 0.4435 & 57.85 & 3.66 & 21.94 & 0.5719 & 0.4547\\
        \rowcolor{gray!15} \cellcolor{white} ~ & SUPIR$^{TSS}$(N=7) & \textbf{4.76} & \textbf{0.5017} & \textbf{58.74} & \textbf{3.96} & \textbf{24.09} & \textbf{0.6736} & \textbf{0.3550} \\
        ~ & PASD(N=7) & 4.99 & 0.5341 & 60.34 & 4.01 & 25.35 & 0.7458 & 0.2762 \\
        \rowcolor{gray!15} \cellcolor{white} ~ & PASD$^{TSS}$(N=7) & \best{\textbf{4.31}} & \textbf{0.5739} & \textbf{62.00} & \best{\textbf{4.08}} & 24.50 & 0.7151 & 0.2767 \\\midrule[1.2pt]
        \multirow{11}{*}{\makecell{DRealSR \\ SR (x4) \\  \quad}} & Real-ESRGAN & 4.35 & 0.5769 & 56.88 & \best{4.34} & 26.26 & 0.7875 & 0.2099\\
        ~ & BSRGAN & 4.60 & 0.6125 & \best{58.55} & 4.31 & \best{27.11} & \best{0.7923} & \best{0.2077} \\
        ~ & SwinIR & 4.39 & 0.5668 & 56.93 & 4.33 & 26.39 & 0.7861 & 0.2082\\\cmidrule{2-9}
        ~ & ResShift (N=15)& 6.03 & 0.6490 & 56.23 & 4.30 & 25.70 & 0.7431 & 0.2863 \\
        ~ & SinSR (N=1)& 5.51 & \best{0.7134} & 56.72 & 4.30 & 25.61 & 0.7426 & 0.2550 \\\cmidrule{2-9}
        ~ & StableSR(N=100) & 4.38 & 0.6472 & 55.88 & 4.26 & 25.07 & 0.7653 & 0.2219 \\
        \rowcolor{gray!15} \cellcolor{white} ~ & StableSR$^{TSS}$(N=100) & 4.73 & \textbf{0.6523} & \textbf{56.12} & \textbf{4.27} & 24.95 & 0.7594 & 0.2224\\
        ~ & SUPIR(N=7) & 5.79 & 0.4760 & 52.69 & 4.09 & 23.04 & 0.5852 & 0.4107 \\
        \rowcolor{gray!15} \cellcolor{white} ~ & SUPIR$^{TSS}$(N=7) & \best{\textbf{4.37}} & \textbf{0.5362} & \textbf{54.26} & \textbf{4.27} & \textbf{25.97} & \textbf{0.7077} & \textbf{0.2910} \\
        ~ & PASD(N=7) & 4.58 & 0.6612 & 58.21 & 4.30 & 26.41 & 0.7754 & 0.2177 \\
        \rowcolor{gray!15} \cellcolor{white} ~ & PASD$^{TSS}$(N=7) & \textbf{3.95} & \textbf{0.6923} & 57.58 & \textbf{4.32} & 25.20 & 0.7327 & 0.2333 \\\midrule[1.2pt]
        ~ & Real-ESRGAN & 3.92 & 0.5709 & 59.25 & 3.64 & - & - & -\\
        ~ & BSRGAN & 5.38 & 0.3305 & 45.46 & 2.11 & - & - & - \\
        \cmidrule{2-9}
        ~ & ResShift (N=15)& 6.59 & 0.6642 & 61.29 & 3.77 & - & - & - \\
        ~ & SinSR (N=1)& 5.91 & \best{0.7610} & 66.43 & 3.90 & - & - & - \\ \cmidrule{2-9}
        RealPhoto60 & StableSR(N=100) & 4.39 & 0.5484 & 55.41 & 3.65 & - & - & - \\
        \rowcolor{gray!15} \cellcolor{white} SR (x2) & StableSR$^{TSS}$(N=100) & \textbf{4.28} & \textbf{0.5575} & \textbf{56.40} & \textbf{3.68} & - & - & - \\
        ~ & SUPIR(N=7) & 5.80 & 0.4597 & 65.42 & 3.62 & - & - & - \\
        \rowcolor{gray!15} \cellcolor{white} ~ & SUPIR$^{TSS}$(N=7) & \best{\textbf{3.86}} & \textbf{0.6277} & \best{\textbf{67.86}} & \best{\textbf{4.38}} & - & - & - \\
        ~ & PASD(N=7) & 4.60 & 0.6244 & 63.85  & 3.99 & - & - & - \\
        \rowcolor{gray!15} \cellcolor{white} ~ & PASD$^{TSS}$(N=7) & \textbf{3.86} & \textbf{0.6427} & \textbf{66.38} & \textbf{4.22} & - & - & - \\\midrule[1.2pt]
        ~ & Real-ESRGAN & 5.92 & 0.4937 & 37.89 & 1.94 & - & - & - \\
        ~ & BSRGAN &  7.16 & 0.4316 & 38.86 & 1.78 & - & - & - \\
        \cmidrule{2-9}
        ~ & ResShift (N=15)& 10.24 & 0.4590 & 29.63 & 1.84 & - & - & - \\
        ~ & SinSR (N=1)& 7.94 & \best{0.6610} & 51.10 & 2.36 & - & - & - \\\cmidrule{2-9}
        WebPhoto & StableSR(N=100) & 7.54 & 0.3412 & 28.03 & 1.72 & - & - & - \\
        \rowcolor{gray!15} \cellcolor{white} SR (x2) & StableSR$^{TSS}$(N=100) & \textbf{7.00} & \textbf{0.3602} & \textbf{28.78} & \textbf{1.73} & - & - & - \\
        ~ & SUPIR(N=7) & 8.56 & 0.3935 & \best{60.42} & 2.95 & - & - & - \\
        \rowcolor{gray!15} \cellcolor{white} ~ & SUPIR$^{TSS}$(N=7) & \best{\textbf{5.19}} & \textbf{0.4815} & 58.72 & \best{\textbf{3.36}}  & - & - & - \\
        ~ & PASD(N=7) & 11.18 & 0.3172 & 28.73 & 1.85 & - & - & - \\
        \rowcolor{gray!15} \cellcolor{white} ~ & PASD$^{TSS}$(N=7) & \textbf{8.57} & \textbf{0.3485} & \textbf{31.25} & \textbf{2.06} & - & - & - \\\midrule[1.2pt]
        ~ & Real-ESRGAN & 5.06 & 0.5337 & 56.60 & 2.75 & - & - & - \\
        ~ & BSRGAN &  6.14 & 0.5566 & 58.48 & 2.60 & - & - & - \\
        \cmidrule{2-9}
        ~ & ResShift (N=15)& 8.42 & 0.5570 & 52.48 & 2.64 & - & - & - \\
        ~ & SinSR (N=1)& 6.97 & \best{0.7553} & 65.40 & 2.98 & - & - & - \\\cmidrule{2-9}
        LFW & StableSR(N=100) & 5.67 & 0.5246 & 55.51 & 3.18 & - & - & - \\
        \rowcolor{gray!15} \cellcolor{white} SR (x2) & StableSR$^{TSS}$(N=100) & \textbf{5.42} & \textbf{0.5416} & \textbf{56.50} & \textbf{3.22} & - & - & - \\
        ~ & SUPIR(N=7) & 6.21 & 0.4329 & 67.09 & 3.50 & - & - & - \\
        \rowcolor{gray!15} \cellcolor{white} ~ & SUPIR$^{TSS}$(N=7) & \best{\textbf{4.36}} & \textbf{0.5776} & \best{\textbf{67.23}} & \best{\textbf{4.22}} & - & - & -  \\
        ~ & PASD(N=7) & 6.51 & 0.5383 & 59.23 & 3.15 & - & - & - \\
        \rowcolor{gray!15} \cellcolor{white} ~ & PASD$^{TSS}$(N=7) & \textbf{5.10} & \textbf{0.5567} & \textbf{62.25} & \textbf{3.55} & - & - & - \\
        \bottomrule[1.2pt]
        
    \end{tabular}
    }
    }
    \caption{Full Quantitative comparison with state-of-the-art Real-world SR methods on synthetic and real-world datasets. `\textbf{Bold}' denotes better than the baseline method. `\underline{Underline}' denotes the best performance. As SwinIR provides pretrained weights only for SR x4, we did not evaluate it on x2 scale factors.}
    \label{tab:sota_real_extra}
\end{table*}

